\documentclass[letterpaper]{article} 
\usepackage{aaai25}  
\usepackage{times}  
\usepackage{helvet}  
\usepackage{courier}  
\usepackage[hyphens]{url}  
\usepackage{graphicx} 
\usepackage{natbib}  
\usepackage{caption} 
\usepackage{algorithm}
\usepackage{algorithmic}
\usepackage{amsmath}
\usepackage{placeins}
\usepackage{booktabs}
\usepackage{xcolor}
\usepackage{amssymb}

\usepackage{newfloat}
\usepackage{listings}
\DeclareCaptionStyle{ruled}{labelfont=normalfont,labelsep=colon,strut=off} 
\floatstyle{ruled}
\newfloat{listing}{tb}{lst}{}
\floatname{listing}{Listing}
\title{ID-Sculpt: ID-aware 3D Head Generation from Single In-the-wild Portrait Image}
\author{
    Jinkun Hao\textsuperscript{\rm 1}, 
    Junshu Tang\textsuperscript{\rm 1}, 
    Jiangning Zhang\textsuperscript{\rm 2},
    Ran Yi\textsuperscript{\rm 1}, 
    Yijia Hong\textsuperscript{\rm 1},
    Moran Li\textsuperscript{\rm 2},\\
    Weijian Cao\textsuperscript{\rm 2},
    Yating Wang\textsuperscript{\rm 1},
    Chengjie Wang\textsuperscript{\rm 2},
    Lizhuang Ma\textsuperscript{\rm 1}\thanks{\textit{Corresponding author.}},
}

\affiliations{
    \textsuperscript{\rm 1}Shanghai Jiao Tong University
    \textsuperscript{\rm 2}Youtu Lab, Tencent

    \{leo\_hao, tangjs, ranyi, hyj542682306, wyating\_0929, lzma\}@sjtu.edu.cn;
    
    \{vtzhang, moranli, weijiancao, jasoncjwang\}@tencent.com;

}

\usepackage{bibentry}

\begin{document}

\maketitle


\begin{abstract}

While recent works have achieved great success on image-to-3D object generation, high quality and fidelity 3D head generation from a single image remains a great challenge.
Previous text-based methods for generating 3D heads were limited by text descriptions and image-based methods struggled to produce high-quality head geometry.
To handle this challenging problem, we propose a novel framework, \textbf{ID-Sculpt}, to generate high-quality 3D heads while preserving their identities. Our work incorporates the identity information of the portrait image into three parts: 1) geometry initialization, 2) geometry sculpting, and 3) texture generation stages. 
Given a reference portrait image, we first align the identity features with text features to realize ID-aware guidance enhancement, which contains the control signals representing the face information. 
We then use the canny map, ID features of the portrait image, and a pre-trained text-to-normal/depth diffusion model to generate ID-aware geometry supervision, and 3D-GAN inversion is employed to generate ID-aware geometry initialization. 
Furthermore, with the ability to inject identity information into 3D head generation, we use ID-aware guidance to calculate ID-aware Score Distillation (ISD) for geometry sculpting. 
For texture generation, we adopt the ID Consistent Texture Inpainting and Refinement which progressively expands the view for texture inpainting to obtain an initialization UV texture map. 
We then use the ID-aware guidance to provide image-level supervision for noisy multi-view images to obtain a refined texture map. Extensive experiments demonstrate that we can generate high-quality 3D heads with accurate geometry and texture from a single in-the-wild portrait image. The project page is at 
\url{https://jinkun-hao.github.io/ID-Sculpt}.
\end{abstract}

\section{Introduction}
\label{sec:intro}

\begin{figure}[t]
    \includegraphics[width=0.45\textwidth]{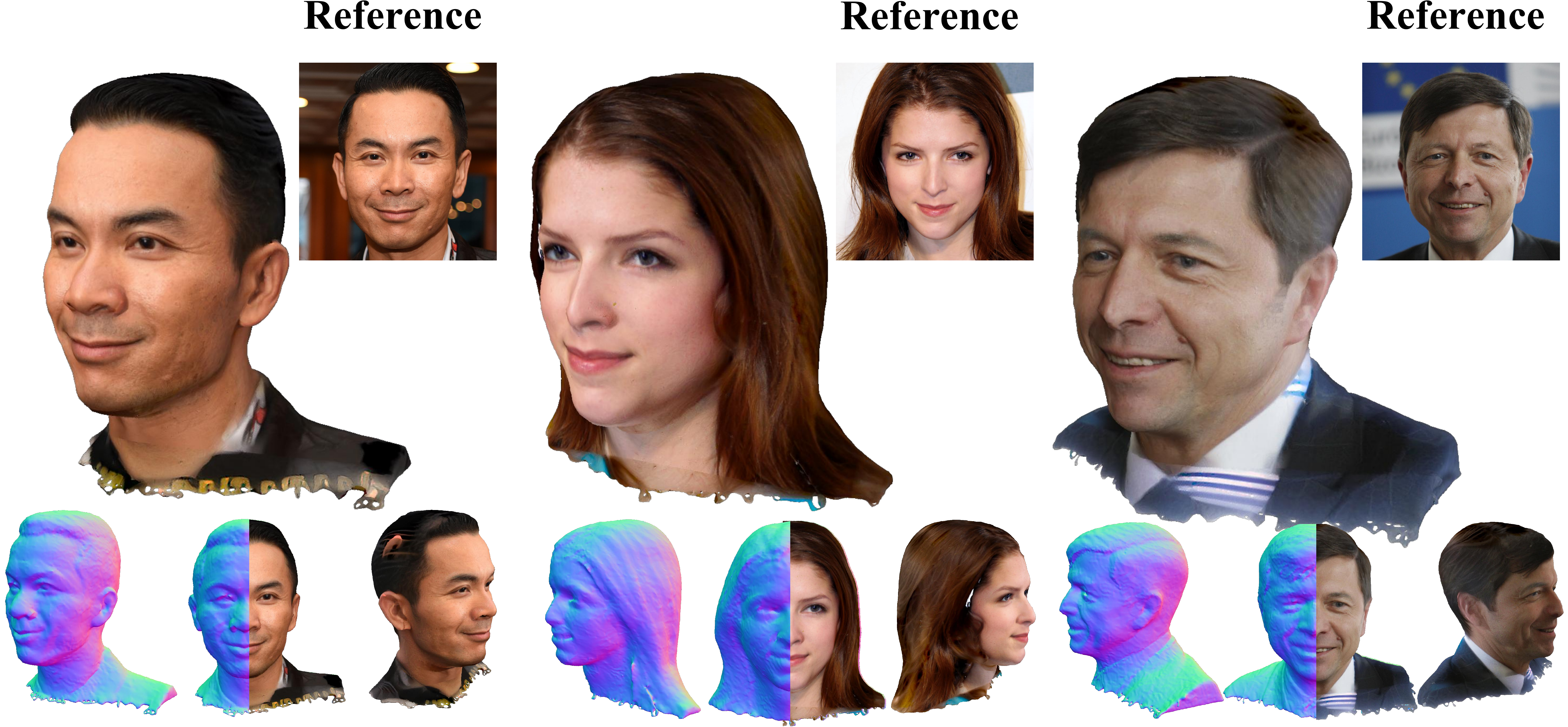} 
    \caption{ID-Sculpt can generate a high-quality 3D head with detailed face geometry and photo-realistic texture given an in-the-wild portrait image.}
    \label{fig:teaser}
\end{figure}

3D head generation aims to create high-quality 3D digital assets of human heads that align with user preferences, utilizing various input data, which has significant applications in film character creation, gaming, online meetings, education, etc.
Conventional methods rely on multi-view geometric consistency~\cite{kirschstein2023nersemble, Zheng_2023_CVPR, munkberg2022extracting, yariv2021volume} or statistical 3D face prior model~\cite{bai2023learning, zheng2023pointavatar, grassal2022neural} to achieve high-quality 3D head generation. The former necessitates multi-view images or several minutes of monocular video, which are not applicable to a single in-the-wild portrait image. Moreover, 3D face shape prior models typically focus on the face region, failing to accurately reconstruct the entire head shape and being unable to handle elements such as long hair and clothing. 
An alternative approach involves employing Generative Adversarial Networks (GANs) to accomplish 3D head generation, which is constrained by limited training data and typically produces frontal-facing faces. Moreover, they struggle to preserve identity consistency when presented with images of large head poses.
Recently, text-to-image~\cite{ramesh2021zero, ramesh2022hierarchical, rombach2022high, saharia2022photorealistic, han2023generalist, pei2024deepfake} generation models trained on large-scale image datasets have achieved significant success. Some approaches employ SDS loss~\cite{poole2022dreamfusion} to ensure similarity between the generated 3D model and images, facilitating image-to-3D~\cite{melas2023realfusion, tang2023make, liu2023syncdreamer, liu2023zero, long2023wonder3d} or text-to-3D~\cite{poole2022dreamfusion, lin2023magic3d, sanghi2022clip, wang2022clip, wang2024prolificdreamer}. However, when employing these methods for customized 3D head generation, the absence of identity information makes it difficult to reproduce the geometric details and texture realism in the portrait image.

\begin{figure}[t]
  \includegraphics[width=0.45\textwidth]{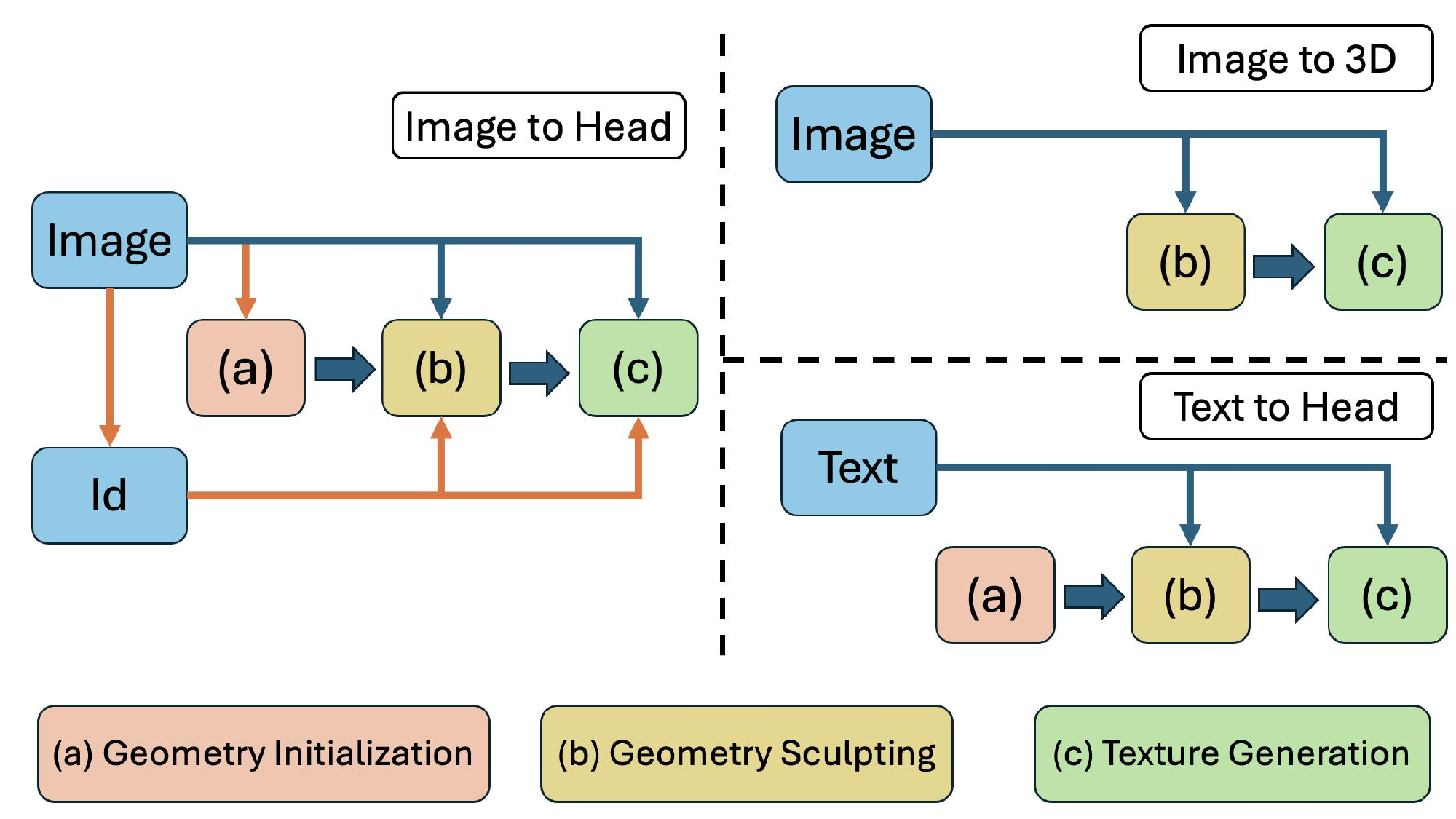}
  \caption{Comparison of previous optimization methods (Right) with our method (Left). Our method leverages the portrait image to get a better initial mesh and utilizes extracted identity information to enhance geometric and texture optimization.
  }
  \label{fig:motivation}
\end{figure}

Through analyzing previous methods, we summarize the key challenges from the three stages in the optimization-based 3D head generation approach:
\textbf{\textit{1)}} Geometry Initialization: Previous image-to-3D methods used an ellipsoid as the initial geometry~\cite{chen2023fantasia3d} and could not accurately reconstruct the geometry of the head. On the other hand, text-to-head methods used a unified FLAME head~\cite{han2024headsculpt, liu2023headartist} as the geometric prior, which lacks personalized information and cannot generate individualized features that deviate from the FLAME head, such as long hair. 
\textbf{\textit{2)}} Geometry Sculpting: Although image-to-3D methods attempted to use view-dependent diffusion ~\cite{liu2023zero}to guide geometry generation, the resulting geometry lacked the generalization ability to the reasonable head geometry. Besides, text-to-3D methods use facial image description text to generate geometry, which can result in a deviation from the identity of the reference image. 
\textbf{\textit{3)}} Texture Generation: Previous methods used VSD~\cite{wang2024prolificdreamer} to mitigate color over-saturation issues based on SDS. However, the generated textures still tend to have excessive shading and unrealistic texture~\cite{huang2023humannorm}.

To address the challenges mentioned above, we incorporate identity information into the three stages of 3D head generation compared with previous methods in Figure~\ref{fig:motivation}. Firstly, we extract Identity embedding using an ID feature extractor and align them with text embedding using a pre-trained ID adapter with decoupled cross-attention. This, along with landmark-guided ControlNet, enables ID diffusion and provides \textbf{ID-aware guidance}, allowing us to incorporate the identity and facial layout information from the reference image into the geometry and texture generation stages. Secondly, we propose \textbf{ID-aware geometry supervision}. By extracting canny maps containing detailed information from the portrait image, we combine canny-guided ControlNet and ID features with off-the-shelf text-to-normal and text-to-depth methods~\cite{huang2023humannorm} to generate high-quality geometric constraint for the geometry generation stage. Additionally, we obtain \textbf{ID-aware initialization} using 3D GAN inversion. In the geometry sculpting stage, we optimize the 3D head using ID-aware guidance with  ID-aware Score Distillation (ISD) loss. In the texture generation stage, we first use Progressive Texture Inpainting to generate a UV-textured map for the uncolored 3D head mesh. Then we leverage ID diffusion to denoise the rendered image from random viewpoints and further refine the texture by calculating the loss between the rendered and denoised images.

In this paper, we propose a new method for generating high-quality 3D heads from single in-the-wild portrait images. The main contributions of this paper are as follows:

\begin{itemize}
  \item A novel pipeline for generating high-quality 3D head models from a single in-the-wild face image.
  \item A method that incorporates identity information into the geometry initialization, geometry generation, and texture generation stages, significantly improving the identity details of the generated 3D heads.
  \item An ID Consistent Texture Inpainting and Refinement method, which uses Progressive Texture Inpainting and refinement method to achieve realistic texture generation.
\end{itemize}

\section{Related Works}
\label{sec:related_work}

\subsection{Text to 3D Generation}

Recent advances in image generative models make it possible to synthesize diverse high-fidelity 3D objects from text prompts. Efforts such as DreamFusion~\cite{poole2022dreamfusion} explore text-to-3D generation by leveraging a Score Distillation Sampling (SDS) loss shows impressive results. Subsequently, Magic3D~\cite{lin2023magic3d} increases the rendering resolution of generated 3D objects. Fantasia3D \cite{chen2023fantasia3d} models the appearance via the BRDF modeling. ProlificDreamer~\cite{wang2024prolificdreamer} proposes Variational Score Distillation (VSD) to mitigate oversaturation problems. Although many works have solved the problem of texture oversaturation to some extent, it is still difficult to generate realistic geometry and texture.

\subsection{Image to 3D Generation}

Image to 3D generation has also developed rapidly. Zero-1-to-3~\cite{liu2023zero} first proposes view-dependent diffusion, which focuses on learning camera viewpoints within ]diffusion models, enabling zero-shot novel view synthesis. GeNVS~\cite{chan2023generative} leverages pixel-aligned features from input views to add 3D awareness to 2D diffusion models. On the other hand, some optimize-based image-to-3D methods, such as Magic123~\cite{qian2023magic123} and Make-it-3D~\cite{tang2023make}, incorporate 3D priors from Zero-1-to-3 and employ coarse-to-fine two-stage optimization. Subsequent works \cite{sun2023dreamcraft3d, yu2023hifi} further refining and enhancing the quality of generated 3D models. However, these methods are limited by the ability of view-dependent diffusion, which cannot accurately generate the geometry from the image.

\subsection{3D Head Generation}

Recently, various approaches~\cite{grassal2022neural, gafni2021dynamic, zheng2023pointavatar, xu2023gaussian} have explored reconstructing 3D head avatars from monocular face videos using different 3D representations. However, these methods necessitate several minutes of video input and struggle to reconstruct less prominent parts in videos. In addition, methods generating heads from 3D GANs like EG3D~\cite{chan2022efficient} proposes a tri-plane representation that can efficiently render high-quality facial images. Panohead~\cite{an2023panohead} advances EG3D by enabling full-head 3D face generation. Rodin~\cite{wang2023rodin} uses diffusion models to generate 3D heads. Nonetheless, they are constrained by the inherent limitations of implicit representations and cannot produce high-quality surface geometry.

Text-based human head generation methods like Headsculpt~\cite{han2024headsculpt} and HeadArtist~\cite{liu2023headartist} integrate facial landmarks to solve layout issues like the Janus face problem. Humannorm~\cite{huang2023humannorm} refines diffusion models using 3D human body data for better normal and depth guidance. Despite these, texture saturation in these methods can impede the attainment of photo-realistic results. In contrast, our method incorporates identity information in all generation stages, which can generate 3D head with great texture realism and identity consistency.

\begin{figure*}[t]
  \includegraphics[width=\textwidth]{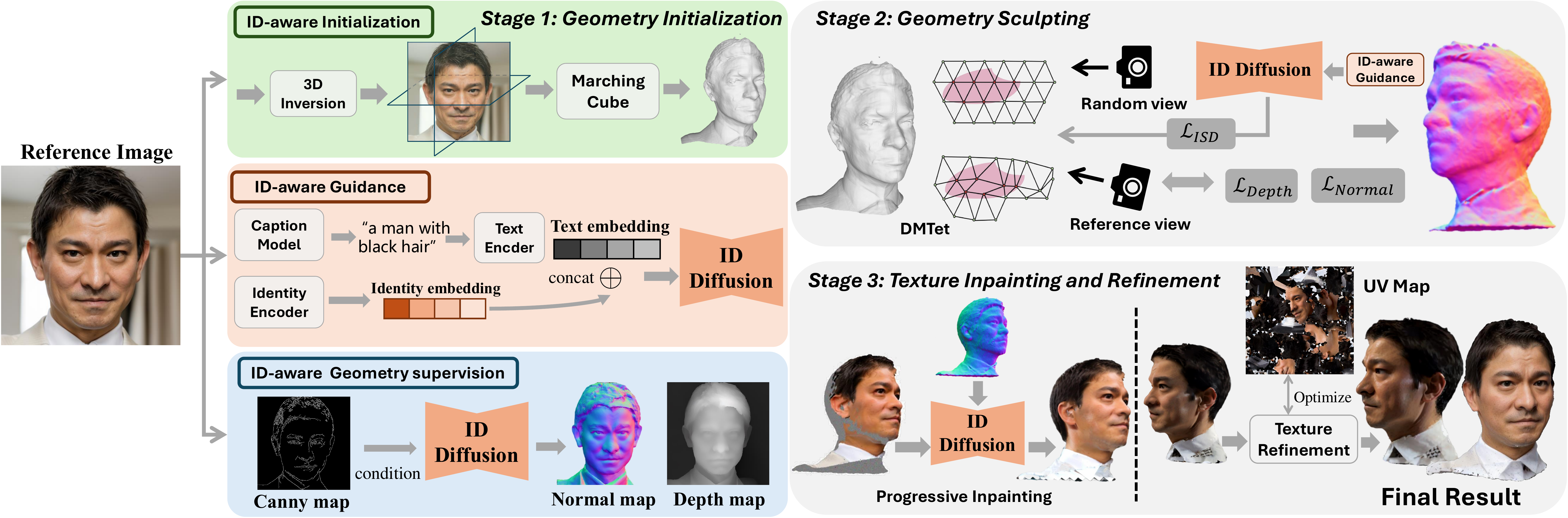}
  \caption{Overview of our proposed ID-Sculpt framework. Given a reference image, ID-Sculpt leverages 3D GAN inversion for improved identity-related geometry initialization in stage 1. In stage 2, ID-aware guidance and ID-aware geometry supervision are used to intergrate identity information into the geometry sculpting process. In stage 3, ID Consistent Texture Inpainting and Refinement is applied to generate a high-quality head texture, where we first use the inpainting method to generate a rough texture and then use image-level ID-aware supervision for texture refinement. With these methods, we can generate high-quality 3D head models with consistent identities from a single in-the-wild face image.}
  \label{fig:pipeline}
\end{figure*}

\section{Method}
In this paper, we propose ID-Sculpt, a novel optimization-based method for high-fidelity 3D head generation from a single portrait image. 
The overall pipeline of ID-Sculpt is shown in Figure~\ref{fig:pipeline}. We fully leverage identity-aware priors to our generation process through \textbf{ID-aware Initialization and Guidance} (§\textcolor{red}{\ref{subsec:ID-aware Initialization and Guidance}}). The generation process consists of two stages: \textbf{Geometry Sculpting} (§\textcolor{red}{\ref{subsec:4.2}}) and \textbf{ID-Consistent Texture Inpainting and Refinement} (§\textcolor{red}{\ref{subsec:4.3}}).

\subsection{ID-aware Initialization and Guidance}
\label{subsec:ID-aware Initialization and Guidance}
Given a facial image $\boldsymbol{x}_{\mathrm{ref}}$ as input, our goal is to generate a high-fidelity 3D head model parameterized with $\theta$, that preserves the identity and appearance of the portrait image.
We adopt an optimization-based pipeline based on the pretrained text-to-image diffusion model $p_t(\boldsymbol{x}_t | y)$, and formulate the optimization process for generating a 3D head as follows:

\begin{equation}
\begin{split}
    &\min \begin{cases}
    \mathbb{E}_{t, c}[\operatorname{D}_{\mathrm{KL}}(q_t^\theta(\boldsymbol{x}_t | \mathbf{c}) \| p_t(\boldsymbol{x}_t | y))], & \mathbf{c} = \mathbf{c}_{\text{rand}} \\
    D(\boldsymbol{x}_{\mathrm{ref}}, \; g(\theta, \boldsymbol{c})), & \mathbf{c} = \mathbf{c}_{\text{ref}}
    \end{cases} \\
    &\text{s.t. } \theta = \{\theta | \boldsymbol{x}_{\mathrm{ref}}, \; \theta_0 \}, \quad \theta_0 = \text{Initial Mesh},
\end{split}
\label{eq:defination}
\end{equation}
where $p_t(\boldsymbol{x}_t | y)$ is the pretrained text-to-image diffusion model and $q_t^\theta(\boldsymbol{x}_t | \mathbf{c})$ is the distribution at time $t$ of the forward diffusion process starting from the rendered image $g(\theta, \boldsymbol{c})$ with the camera $\mathbf{c}$. $\mathbf{c}_{\text{ref}}$ is the reference camera pose. 
$D(\cdot)$ denotes the distance function, and $\theta_0$ is the initial 3D mesh provided by users.

For 3D full head generation task, the direct description of the input facial image, often cannot fully capture the variations of different facial features. Besides, pretrained 2D view-dependent diffusion models (\textit{e.g.,} Zero-123~\cite{liu2023zero}) are trained on general objects, which cannot generalize to human heads. Therefore, our key insight is to leverage identity knowledge and inject state-of-the-art identity priors into the mesh initialization and diffusion guidance.

\noindent\textbf{ID-aware Initialization.}
The initialization is crucial for 3D generation.
According to the optimization goal in Eq.(\ref{eq:defination}), a good initial 3D mesh $\theta_0$ should possess the approximate geometric shape of the face in the given image $\boldsymbol{x}_{\mathrm{ref}}$. 
Some approach uses FLAME~\cite{DECA:Siggraph2021} to obtain the initial 3D head shape. 
However, this approach cannot generate personalized facial features and does not describe the geometric shape of hair. 
To address this issue, we employ a 3D-aware GAN model~\cite{an2023panohead} 
trained on a comprehensive $360^\circ$ head dataset  
to obtain the initial 3D head mesh $\theta_0$. Specifically,
given the input facial image $\boldsymbol{x}_{\mathrm{ref}}$, 
we first perform inversion to find latent code $z$ and then perform Pivotal Tuning Inversion (PTI)~\cite{roich2022pivotal} to obtain a 3D representation that better matches the input image.
Through 3D GAN inversion, we can obtain an initial  
3D head mesh that corresponds to the input facial image, denoted as:
\begin{equation}
  \theta_0 = \text{3D-GAN}(z_0 | \boldsymbol{x}_{\text{ref}}).
\end{equation}

\noindent\textbf{ID-aware Guidance.}
In order to obtain finer information about the facial image, we  
extract high-dimensional identity features $I_{r}$ from the reference image 
as additional conditions beyond text prompt, to guide the denoising process of diffusion models.
We use a facial recognition model~\cite{deng2019arcface} trained on large-scale facial datasets as the identity encoder to extract robust identity features. 
Additionally, we utilize 
a light projection network to map the extracted ID feature
to an enhanced ID feature $f_{ID}$ aligned with text embedding, and 
insert the ID feature into diffusion
with decoupled cross-attention~\cite{ye2023ip}.
In this way, we construct an ID-guided diffusion model $p^{\text{id}}$, which will be utilized in our ID-aware Score Distillation loss to provide identity  
guidance for a specific person.

Identity-enhanced prompt only extracts general identity information, but cannot provide the information of 
3D facial layout. 
For example, a person has different mouth shapes when laughing and crying, but  
the identity features are the same.
To address this issue, we utilize a pretrained facial shape predictor~\cite{guo2020towards} 
to obtain 
3D facial landmarks 
of the reference image  
$\boldsymbol{x}_{\mathrm{ref}}$,
and align the 3D facial landmarks to
the initial 3D head mesh. 

Given a camera pose $\boldsymbol{c}$, we project the extracted 3D facial landmarks onto the 2D plane, and obtain 
a facial landmark image $L(\mathbf{c})$, 
which provides the facial component layout information from that viewpoint.
We then apply a landmark-guided ControlNet~\cite{zhang2023adding} to introduce the facial 
layout information into diffusion guidance.

By incorporating three types of information, \textit{i.e.,} text prompt $y$ (obtained from $\boldsymbol{x}_{\mathrm{ref}}$ by BLIP-2~\cite{li2023blip}), identity feature $f_{ID}$, and the projected facial landmark image $L(\mathbf{c})$, we propose \textbf{ID-aware Score Distillation (ISD)} loss, which can be formulated as:
\begin{equation}
\mathcal{L}_{\mathrm{ISD}}=\mathbb{E}_{t, c}[D_{\mathrm{KL}}(q_t^\theta(\boldsymbol{x}_t | \mathbf{c}) \| p^{\text{id}}_t(\boldsymbol{x}_t | y, f_{ID}, L(\mathbf{c}), \mathbf{c}))].
  \label{eq:ISD}
\end{equation}

Compared to SDS, ISD introduces more ID-preserved supervision that is aligned with the reference portrait image.

\begin{figure}[t]
  \includegraphics[width=0.45\textwidth]{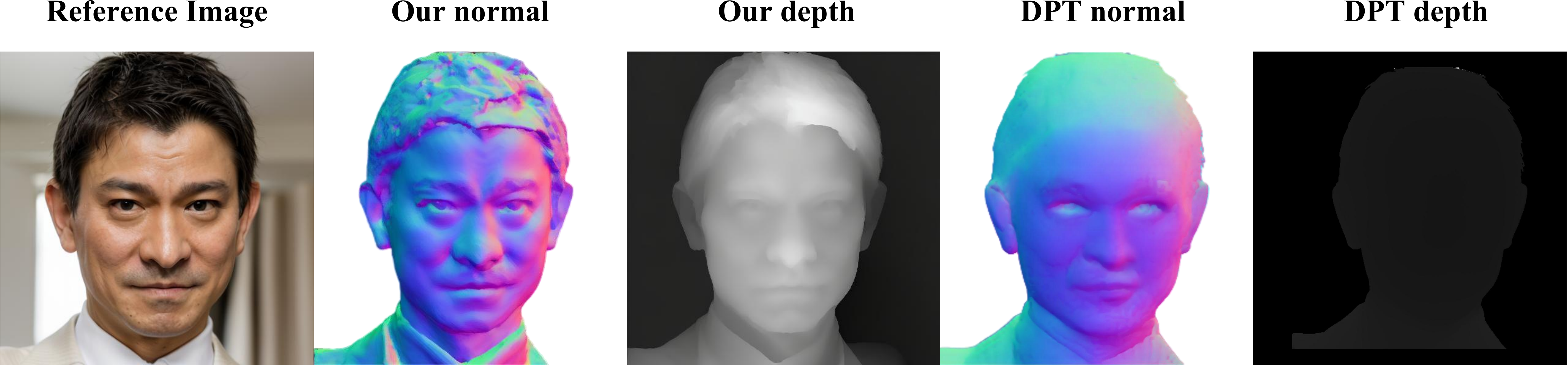}
  \caption{Comparison between our normal map and depth map and those generated by DPT\cite{Ranftl2021, Ranftl2020}.}
  \label{fig:png}
\end{figure}

\noindent\textbf{ID-aware Geometry Supervision}. 
In the geometry sculpting stage, the reference image $\boldsymbol{x}_{\mathrm{ref}}$ in RGB space cannot be directly used 
as the supervision for geometry.
A common method is to use normal maps and depth maps that contain geometric information for supervision, which can be predicted based on the reference image in RGB space. 
Therefore, it is crucial to obtain high-quality geometric representations $\boldsymbol{X}_{\text{ref}}^{\text{geo}}$ from the reference image in RGB space. 
Previous methods~\cite{eftekhar2021omnidata} directly predicted depth maps and normal maps from images but failed to capture some geometric details in the reference images, which could result in flat geometries, as shown in Fig~\ref{fig:png}. 
Different from previous methods, we first use an edge extractor to extract high-frequency facial geometric details and obtain a Canny image as the condition. 
Then we utilize pre-trained Text-to-Normal and Text-to-Depth diffusion models~\cite{huang2023humannorm} with ControlNet to generate normal maps and depth maps. 
Additionally, to better preserve identity information, we also add ID features into the diffusion process. 
Through these approaches, high-quality geometric supervision in the reference view can be formulated as:
\begin{equation}
  \boldsymbol{X}_{\text{ref}}^{\text{geo}} = p^{\text{id}}_{\text{geo}}(\boldsymbol{x}_t | y, f_{\text{ID}}, \text{Canny}(\boldsymbol{x}_{\text{ref}})),
\end{equation}
where $\boldsymbol{X}_{\text{ref}}^{\text{geo}} = \{ \boldsymbol{X}_{\text{ref}}^{\text{normal}}, \boldsymbol{X}_{\text{ref}}^{\text{depth}} \} $ and $p^{\text{id}}_{\text{geo}} = \{ p^{\text{id}}_{\text{normal}}, p^{\text{id}}_{\text{depth}} \}$. In this way, the 3D head model can obtain geometric details that are as rich as the information in the RGB image of the reference view.

\subsection{Geometry Sculpting}
\label{subsec:4.2}

During the geometry sculpting stage, we aim to obtain a fine-grained 3D geometry, that is consistent with the input image under the reference view, and consistent across different viewpoints. 
We first convert the initial 3D mesh obtained from ID-aware initialization into DMTet~\cite{shen2021dmtet} representation to facilitate high-resolution details and high-quality surface. 
Specifically, we use the normals rendered from 3D-GAN as the pseudo ground truth normals $\mathcal{N}_{\mathbf{c}}$. 
For the DMTet object to be optimized, we use Marching Tetrahedra to generate the normal map $\hat{\mathcal{N}_{\mathbf{c}}}$. 
To polish the overall surface from multiple viewpoints, we uniformly sample camera poses $\mathbf{c}$ distributed in the 3D-GAN space, and use $L_2$ distance between normal maps as the loss function, defined as $\mathcal{L}_\text{norm} = \| \hat{\mathcal{N}_{\mathbf{c}}} - \mathcal{N}_{\mathbf{c}}\|_2$.

\noindent\textbf{Geometry Refinement.}
During the geometry refinement stage, we use \textbf{ID-aware Geometry Supervision} to ensure the generation of fine geometry under the reference view, while supplementing the information with the prior of the diffusion model under other views, as shown in Eq.(\ref{eq:defination}). 
Under the reference view, we calculate the reference loss from the 
head foreground mask, normal map, and depth map of the reference image, which is defined as:
\begin{equation}
    \mathcal{L}_{\text{ref}} = \mathcal{L}_{\text{mask}} + \mathcal{L}_{\text{normal}} + \mathcal{L}_{\text{depth}},
\end{equation}
where the foreground mask $M$ is obtained through face parsing, and both the mask loss $L_{\text{mask}}$ and normal loss $L_{\text{normal}}$ are defined as the $\mathcal{L}_2$ distance between the rendered mask and normal map, and the reference mask and normal map. 
For the depth loss $L_{\text{depth}}$, we use the negative Pearson correlation to alleviate the problem of depth scale mismatch.

For other views that are randomly sampled around the 3D head, we use the ISD loss $L_{\mathrm{ISD}}$ in Eq.(\ref{eq:ISD}) to guide the optimization of the geometry, based on the ID-enhanced diffusion prior. 
Finally, the Marching Tetrahedra algorithm is applied to extract the mesh.

\subsection{ID-Consistent Texture Inpainting and Refinement}
\label{subsec:4.3} 
During the texture generation stage, our goal is to rapidly generate realistic textures for the 3D head mesh. 
Previous textures generated from SDS/VSD loss~\cite{chen2023fantasia3d,wang2024prolificdreamer} may become over-saturated and inconsistent with the color tone of the reference image.
Therefore, we design an inpainting and refinement process to quickly produce high-quality UV texture maps.

\noindent \textbf{Progressive Texture Inpainting.}
With an uncolored 3D head surface mesh $M$ with vertices $V$ and triangular faces $F$ generated in the geometry generation stage, denoted as $M = (V,F)$, we initialize a UV texture map $\mathbf{T}_0$ corresponding to the mesh coordinates. 
Starting from the reference view $\mathbf{c}_0$, we first back-project the reference portrait image onto the 3D mesh. To maintain consistency with the reference image texture, we sample a camera trajectory  
$\{\mathbf{c}_1, \mathbf{c}_2, \cdots, \mathbf{c}_n\}$ (where $\mathbf{c}_i$ is the $i$-th viewpoint, and $n$ is the number of sampled viewpoints)
to progressively generate texture. The camera trajectory starts from the reference image viewpoint, gradually moves the camera pose to the back view of the head, and finally generates the texture for the top of the head that has not been generated yet. 

During the inpainting process, at a specific viewpoint $\mathbf{c}_k$, given the texture map $\mathbf{T}_{k-1}$ that has been textured in $k-1$ previous viewpoints, we can render an RGB image $I_k$ with partially colored regions, as well as a mask $\mathbf{m}_{k-1}$ indicating the  
colored (visible)
regions.
To better inpaint the uncolored areas, we render the normal map $\boldsymbol{x}^{normal}_k$ from the 3D mesh as the geometric condition at the $\mathbf{c}_k$ viewpoint, and then use the normal-conditioned diffusion model to generate the texture for the uncolored areas, which is formulated as:
\begin{equation}
    \hat{I}_k = p^{\text{id}}_0(\boldsymbol{x}_0 | y, f_{ID}, \boldsymbol{x}^{normal}_k).
\end{equation}

The generated image $\hat{I}_k$ at the viewpoint $\mathbf{c}_k$ will be projected back onto the 3D mesh, resulting in the inpainted texture map $\hat{\mathbf{T}}_k$. 
And the uncolored region ($1 - \mathbf{m}_{k-1}$) in UV texture $\mathbf{T}_{k-1}$ is updated 
as follows:
\begin{equation}
    \mathbf{T}_{k} = \mathbf{m}_{k-1} \odot \mathbf{T}_{k-1} + (1 - \mathbf{m}_{k-1}) \odot \hat{\mathbf{T}}_k,
\end{equation}
where $\mathbf{T}_{k}$ is the updated UV texture map after $k$ step inpainting. With this progressive inpainting method, we can gradually inpaint the textures from other viewpoints starting from the reference portrait image.

\begin{figure*}[t]
  \includegraphics[width=\textwidth]{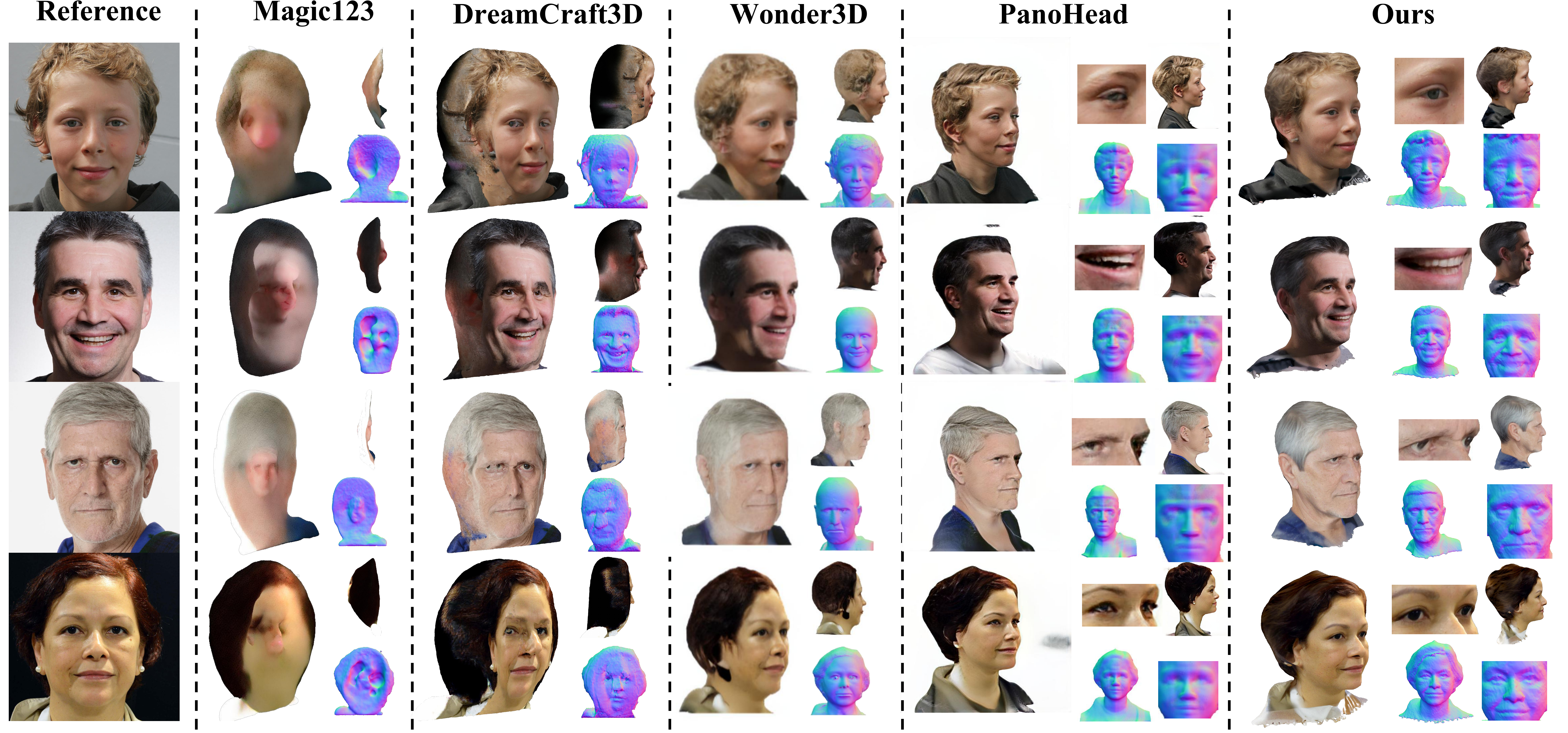}
  \caption{Qualitative evaluation with image-to-3D methods. Our method surpasses previous approaches in terms of the facial detail in head geometry as well as the realism of the texture. The zoom-in result of eye/mouth region and facial geometry shows that our method has clearer structural details and more reasonable geometric structure.
  }
  \label{fig:image-based comparation}
\end{figure*}

\noindent \textbf{Texture Refinement.}
In the texture refinement stage, we aim to solve the texture inconsistency problem that appears in texture inpainting and further enhance the detail and realism of the generated texture. 

We optimize the UV texture map by utilizing the ID-guided diffusion model and minimizing the distance between the rendered images and the refined images.

Specifically, for 
a rendered image $\boldsymbol{x}_0$ from a sampled viewpoint, we add Gaussian noise with $t$ diffusion steps and use the ID-guided diffusion model $p^{id}$ to restore it. 
Moreover, we add rendered normals $\boldsymbol{x}^{normal}$ from the corresponding viewpoint as additional geometric information conditions during the diffusion denoising process.
This yields a refined image $\hat{\boldsymbol{x}}_0$ in the respective viewpoint which we will use as supervision to optimize the texture map:
\begin{equation}
    \hat{\boldsymbol{x}}_0 = p^{\text{id}}_0(\boldsymbol{x}_0 | \boldsymbol{x}_t, y, f_{ID}, \boldsymbol{x}^{normal}).
\end{equation}

After obtaining the refined image $\hat{\boldsymbol{x}}_0$, we calculate the pixel-level MSE loss $\mathcal{L}_{\text{MSE}}$ between $\boldsymbol{x}_{0}$ and $\hat{\boldsymbol{x}}_0$ to further refine the texture map. 
Besides, we add a perceptual loss $\mathcal{L}_{\text{percep}}$ to enhance the texture detail information and maintain style similarity. 
We also calculate the MSE loss  $\mathcal{L}_{\text{MSE}}^{\text{ref}}$ and perceptual loss $\mathcal{L}_{\text{percep}}^{\text{ref}}$ between $x_0$ from the reference view and the reference image $\boldsymbol{x}_{\mathrm{ref}}$. 
By combining these supervisions, we obtain the final loss $\mathcal{L}_{\text{refine}}$ for the texture refinement stage:
\begin{equation}
    \mathcal{L}_{\text{refine}} = \mathcal{L}_{\text{MSE}} + \mathcal{L}_{\text{percep}} + \mathcal{L}_{\text{MSE}}^{\text{ref}} + \mathcal{L}_{\text{percep}}^{\text{ref}}.
\end{equation}

\section{Experiments}

In the experimental section, we start with the implementation details, followed by quantitative and qualitative comparisons between our method and others. Finally, we analyze the role of each module in our method.

\subsection{Implementation Details}

Our ID-Sculpt is built upon the open-source project ThreeStudio~\cite{threestudio2023}. In the geometry stage, we use SD1.5~\cite{rombach2022high} as the base diffusion model. During the texture generation stage, we employ Realistic Vision 4.0 as the base diffusion model. 
In the geometry generation stage, we perform 5,000 optimization iterations. During the texture generation stage, we first perform texture inpainting from 15 ordered viewpoints, followed by 400 steps of texture refinement. Our experiments are conducted on a single V100 GPU, with a batch size set to 1. For each 3D head, the total optimization time is approximately 40 minutes.

\noindent\textbf{Baselines.} We conducted extensive comparisons with four other single-view 3D generation methods (Magic123~\cite{qian2023magic123}, DreamCraft3D\cite{sun2023dreamcraft3d}, Wonder3D~\cite{long2023wonder3d}) and a single-view human head generation method (Panohead~\cite{an2023panohead}). For Magic123, DreamCraft3D, and Wonder3D, we generated the 3D models strictly following the official open-source code and parameters. For Panohead, we followed the official guidance and obtained results using the GAN inversion method. In addition, we compared our method with the state-of-the-art text-to-3D human head generation methods HumanNorm~\cite{huang2023humannorm}, for it has excellent generation results and good reproducibility.

\noindent\textbf{Datasets.} We conducted experiments on a subset of facial images from the FFHQ~\cite{karras2019style} dataset, a high-quality in-the-wild facial dataset known for its rich diversity in age, ethnicity, and image backgrounds, as well as significant variations in facial attributes. This dataset presents a challenging task due to its unconstrained environment. From this dataset, We selected 106 portrait images with unoccluded faces to evaluate our model and baselines. 

\noindent\textbf{Evaluation metrics.} We employ LPIPS~\cite{zhang2018unreasonable} for fidelity measurement, CLIP-I~\cite{ye2023ip} score and ID~\cite{an2023panohead} score to measure the multi-view consistency and identity preservation.

\begin{figure}[t]
  \includegraphics[width=0.45\textwidth]{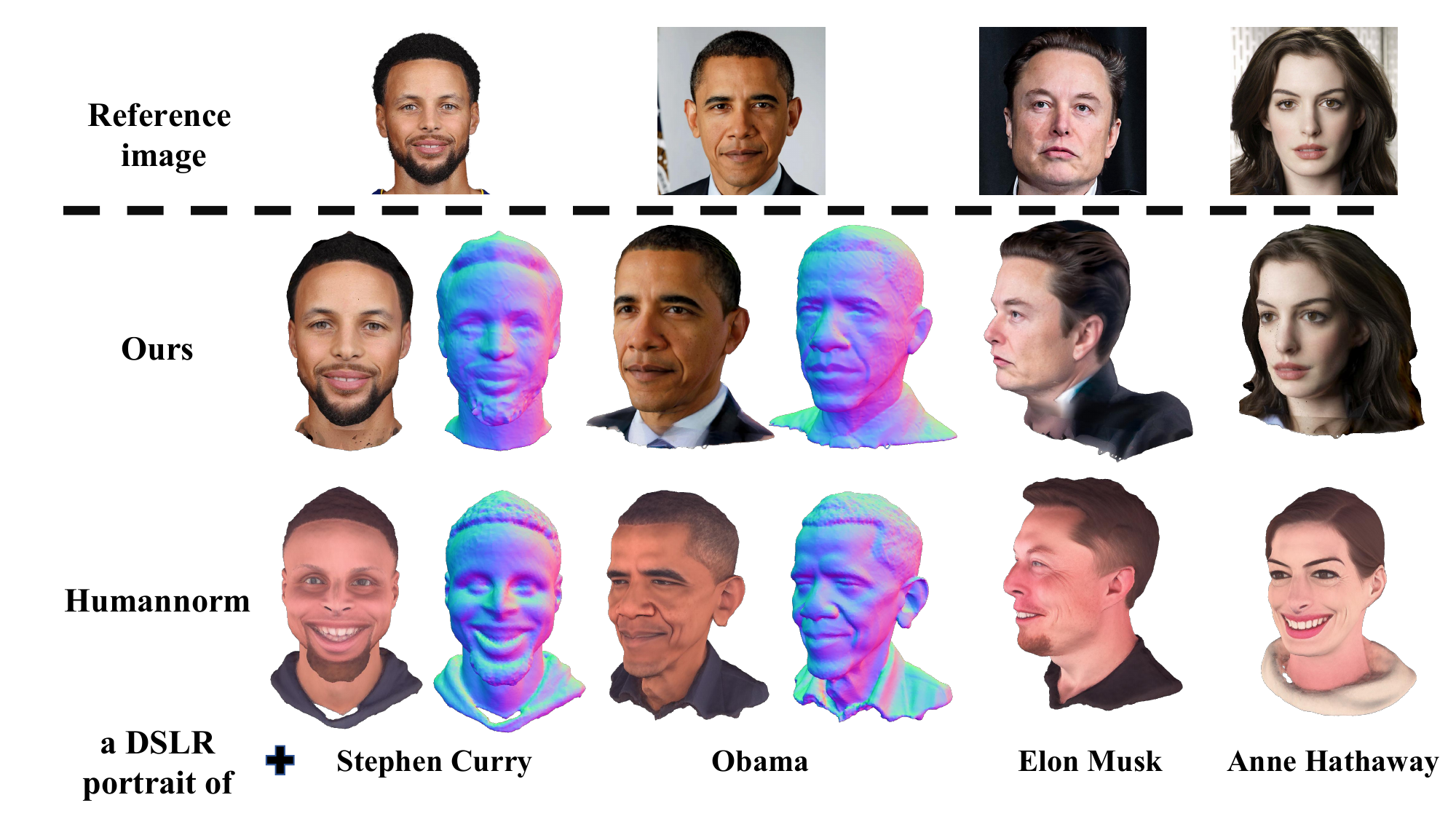}
  \caption{Qualitative evaluation with text-to-head method. Our method has a clear advantage in the realism.
  }
  \label{fig:text-based comparation}
\end{figure}

\subsection{Qualitative Evaluation}

\textbf{Qualitative evaluation with image-to-3D methods.} In Fig.~\ref{fig:image-based comparation} we show the comparisons between our method and four other image-to-3D methods. These four images are selected from the in-the-wild face image dataset FFHQ and include faces of different ages and genders. For the geometry, the 3D generation methods based on optimization (Magic123 and Dreamcraft3D) can't restore the stereoscopic shape of the human head. Wonder3D can generate basic human head shapes but generally suffers from overly flat facial features in side views, and the facial surface lacks geometry detail. Besides, Panohead produces strange protrusions in the eye area. In contrast, our method, after introducing the identity information to the geometry sculpting stage, can restore fine geometry on the face while ensuring a high-quality overall head shape. For texture results, Dreamcraft3D and Wonder3D cannot generate images of the head from other angles. Wonder3D's texture quality drops significantly on the side (i.e., the side rear view of the head in the first row of Wonder3D) and cannot produce high-resolution textures. Panohead struggles to generate high-fidelity structural details in areas like the eye and mouth region. After using ID-enhanced inpainting and refinement processes, our method can generate more photorealistic textures. In summary, our method can produce high-quality head geometry and realistic textures while fully restoring the facial information from the reference images.

\noindent\textbf{Qualitative evaluation with text-to-head method.} In Fig.~\ref{fig:text-based comparation}, we show a comparison of our method with the current SOTA text-to-head method HumanNorm. Since text description cannot fully describe in-the-wild face information, we choose examples of celebrities so that the text-to-head method can accurately generate heads based on the celebrity names. For generated geometry, HumanNorm tends to generate heads with exaggerated structures, whereas our method generates far more realistic geometries. In terms of texture, the heads generated by HumanNorm have over-saturated colors. In contrast, our method fully utilizes both the reference image information and the identity information of the person depicted to generate more realistic faces.

\subsection{Quantitative Evaluation}

\begin{table}
\centering
\small
\setlength{\tabcolsep}{4pt} 
  \caption{Quantitative comparisons on head generation task. We compute LPIPS under the reference view, ClIP-I and ID score under novel views.}
  \label{tab:quanti}
  \begin{tabular}{lcccc}
    \toprule
    Methods & LPIPS $\downarrow$ & CLIP-I $\uparrow$ & ID $\uparrow$ & User Study $\uparrow$\\ \hline
    Magic123 & 0.367 & 0.4447 & 0.0245 & 1.11\\
    PanoHead & \underline{0.045} & 0.6600 & \underline{0.2737} & \underline{4.09}\\
    Wonder3D & 0.135 & \textbf{0.6860} & 0.2729 & 3.13\\
    DreamCraft3D & 0.068 & 0.6074 & 0.2526 & 1.95\\
    ID-Sculpt (Ours) & \textbf{0.039} & \underline{0.6605} & \textbf{0.4451} & \textbf{4.72}\\
  \bottomrule
  \end{tabular}
\end{table}

For each generated head model, we render images from four consistent viewpoints: left-frontal, left, right-frontal, and right. In Tab. \ref{tab:quanti}, we present the quantitative comparisons with the previous SOTA methods. Our method outperforms previous approaches in terms of the ID metric and LPIPS score, indicating its effectiveness in capturing facial features and demonstrating high identity consistency across various perspectives in the analysis of novel viewpoints.

\noindent\textbf{User Study.} To validate our model's robustness and quality, we conducted a user study with 15 random examples. Participants ranked five methods' outputs based on quality from two random viewpoints. Our model ranked highest on average across 480 responses from 32 participants.

\subsection{Ablation Study}
\begin{figure}[t]
  \includegraphics[width=0.45\textwidth]{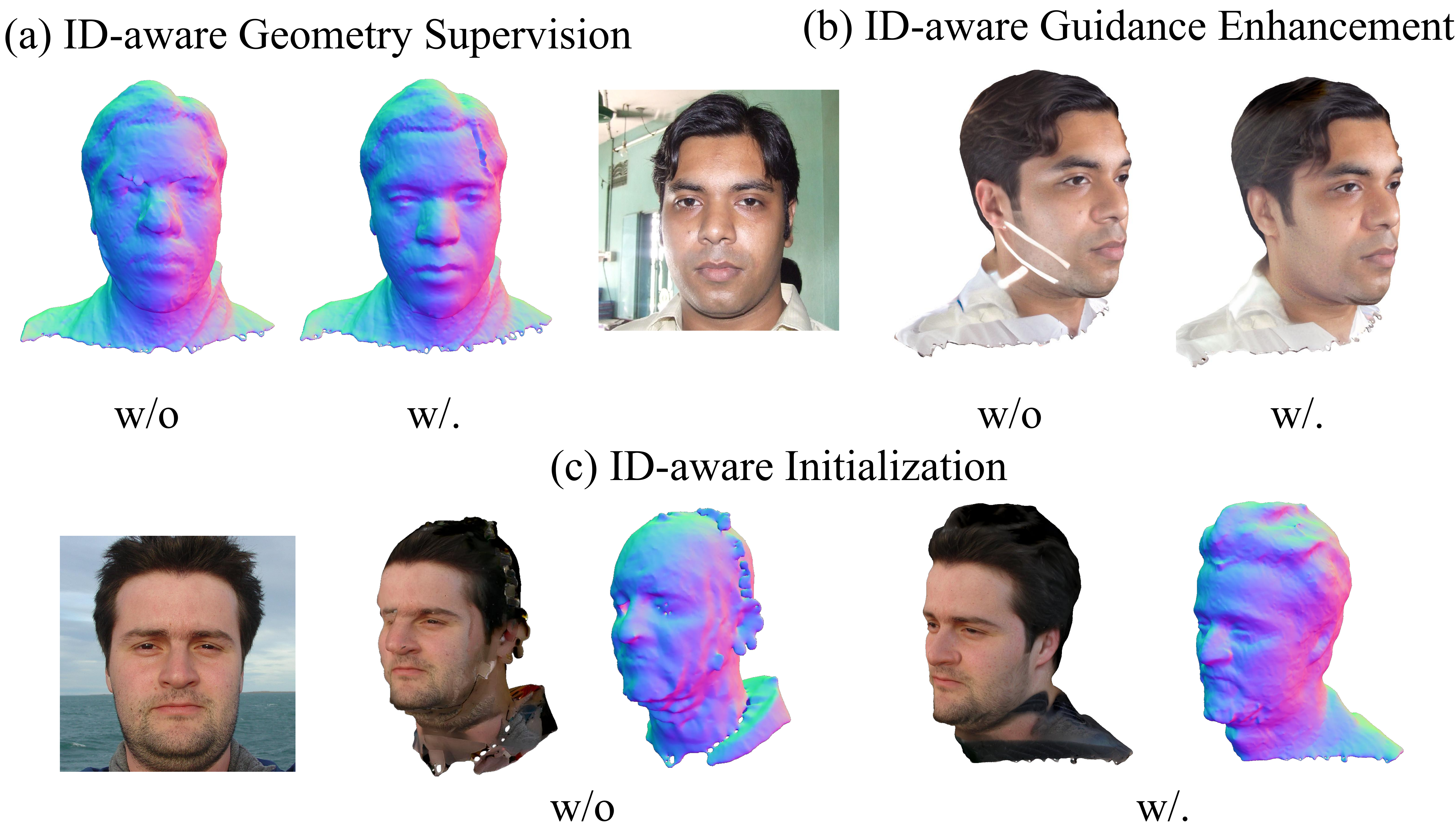}
  \caption{Visual ablation results of our ID-aware Initialization and Guidance. 
  }
  \label{fig:ID-aware ablation}
\end{figure}

\begin{figure}[t]
  \includegraphics[width=0.45\textwidth]{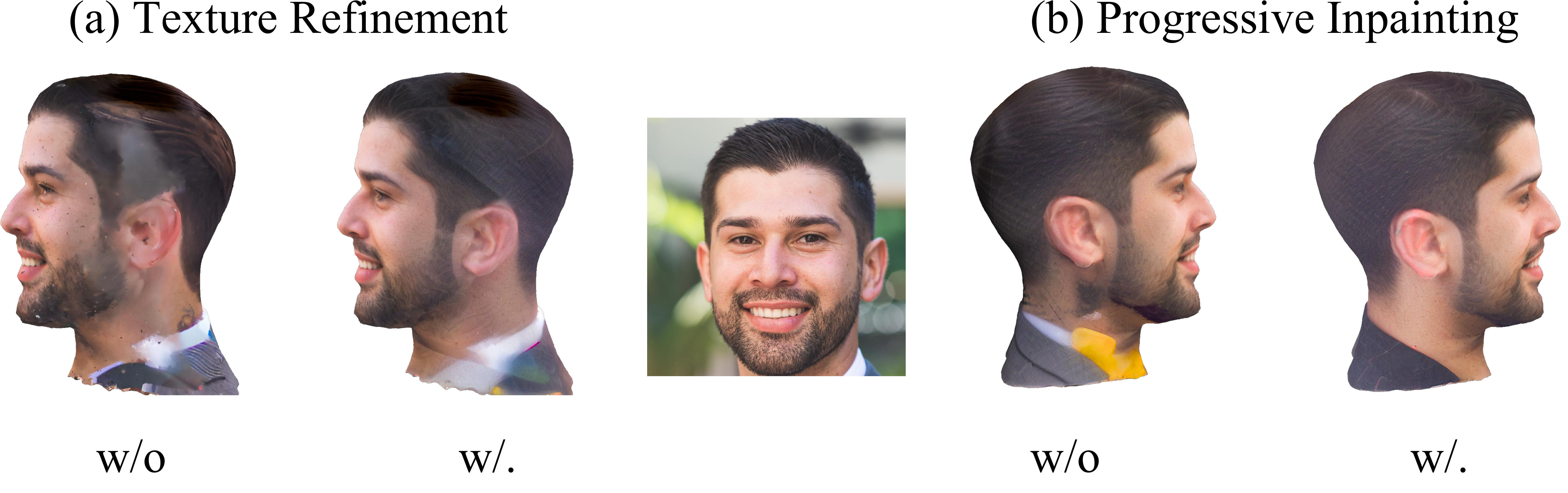}
  \caption{Visual ablation results of our ID-Consistent Texture Inpainting and
Refinement. Our method effectively reduces the artifacts in texture generation.
  }
  \label{fig:ablation texture}
\end{figure}

\begin{figure}[t]
  \includegraphics[width=0.45\textwidth]{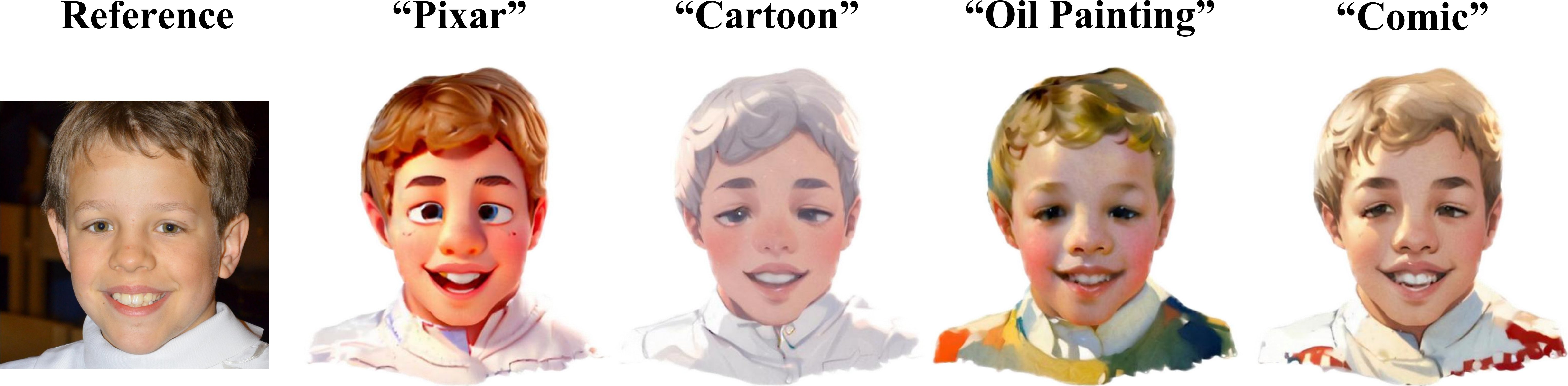}
  \caption{3D head with different styles. With the stylized diffusion model, we can achieve diversified stylized 3D head generation while maintaining identity.
  }
  \label{fig:stylized texture}
\end{figure}

\textbf{Effectiveness of ID-aware Initialization and Guidance.} We present the visual result in Fig.\ref{fig:ID-aware ablation} result to show the effectiveness of our ID-aware Initialization and Guidance method. (a) By replacing the normal map obtained through ID-aware Geometry Supervision with the normal map extracted from DPT\cite{Ranftl2021, Ranftl2020}, we can observe that without ID-aware Geometry Supervision, the generated geometry loses many facial details. (b) When removing the ID-aware Guidance Enhancement in the texture generation stage, we can see that the faces lose the constraints imposed by identity information and easily produce artifacts for large angles (e.g., side faces) without sufficient reference image information. (c) We compare our ID-aware Initialization results by using Flame as the initialization geometry. As Flame aligns only with the facial landmarks, the head portions other than the face in the images are difficult to align with the initialization geometry, leading to difficulty in geometry optimization.

\noindent\textbf{Effectiveness of ID-Consistent Texture Inpainting and Refinement.} As shown in Fig.\ref{fig:ablation texture}, without (a) Texture refinement=, relying solely on inpainting can lead to discontinuous boundaries between each inpainting step, which is particularly noticeable at the intersection of the ears and hair. Without (b) Progressive Inpainting to complete the texture and instead use a view-surrounding approach, incoherent texture colors will be generated (e.g. yellow part at the neck).

\subsection{Application}

\textbf{ID-consistant Texture Stylization.} With the capability of stylized text-to-image diffusion models, we achieve stylized 3D head generation. Due to the use of ID-aware guidance in the texture generation stage, the generated stylized heads can maintain the identity information of the character in the portrait image, as shown in Fig~\ref{fig:stylized texture}.

\section{Conclusion}
We propose ID-Sculpt, a novel method to generate a 3D head from a single in-the-wild portrait image. Our method significantly improves ID consistency and geometric details by adding identity knowledge to the initialization, guidance, and geometry supervision part of the optimization process. Moreover, we design a two-stage ID-consistent texture inpainting and refinement method, which leads to coherent and photo-realistic head texture. ID-Sculpt outperforms previous methods in both geometry and texture quality, demonstrating the ability to generate high-fidelity 3D heads. These advantages combined with stylized 3D head generation bring great help for personalized 3D head asset generation.

\noindent\textbf{Limitations and future work.} Although we can generate high-quality 3D heads from a portrait image, the optimization-based approach takes the whole generation a long time (40 minutes). In the future, we will work on accelerating the speed of 3D head generation.

\section*{Acknowledgments}
This work was supported by National Natural Science Foundation of China (No. 72192821, 62302297, 62472282), Shanghai Sailing Program (22YF1420300), Young Elite Scientists Sponsorship Program by CAST (2022QNRC001), the Fundamental Research Funds for the Central Universities (project number: YG2023QNA35).

\bibliography{aaai25}

\begin{thebibliography}{51}
\providecommand{\natexlab}[1]{#1}

\bibitem[{An et~al.(2023)An, Xu, Shi, Song, Ogras, and Luo}]{an2023panohead}
An, S.; Xu, H.; Shi, Y.; Song, G.; Ogras, U.~Y.; and Luo, L. 2023.
\newblock Panohead: Geometry-aware 3d full-head synthesis in 360deg.
\newblock In \emph{Proceedings of the IEEE/CVF Conference on Computer Vision and Pattern Recognition}, 20950--20959.

\bibitem[{Bai et~al.(2023)Bai, Tan, Huang, Sarkar, Tang, Qiu, Meka, Du, Dou, Orts-Escolano et~al.}]{bai2023learning}
Bai, Z.; Tan, F.; Huang, Z.; Sarkar, K.; Tang, D.; Qiu, D.; Meka, A.; Du, R.; Dou, M.; Orts-Escolano, S.; et~al. 2023.
\newblock Learning Personalized High Quality Volumetric Head Avatars from Monocular RGB Videos.
\newblock In \emph{Proceedings of the IEEE/CVF Conference on Computer Vision and Pattern Recognition}, 16890--16900.

\bibitem[{Chan et~al.(2022)Chan, Lin, Chan, Nagano, Pan, De~Mello, Gallo, Guibas, Tremblay, Khamis et~al.}]{chan2022efficient}
Chan, E.~R.; Lin, C.~Z.; Chan, M.~A.; Nagano, K.; Pan, B.; De~Mello, S.; Gallo, O.; Guibas, L.~J.; Tremblay, J.; Khamis, S.; et~al. 2022.
\newblock Efficient geometry-aware 3d generative adversarial networks.
\newblock In \emph{Proceedings of the IEEE/CVF conference on computer vision and pattern recognition}, 16123--16133.

\bibitem[{Chan et~al.(2023)Chan, Nagano, Chan, Bergman, Park, Levy, Aittala, De~Mello, Karras, and Wetzstein}]{chan2023generative}
Chan, E.~R.; Nagano, K.; Chan, M.~A.; Bergman, A.~W.; Park, J.~J.; Levy, A.; Aittala, M.; De~Mello, S.; Karras, T.; and Wetzstein, G. 2023.
\newblock Generative novel view synthesis with 3d-aware diffusion models.
\newblock In \emph{Proceedings of the IEEE/CVF International Conference on Computer Vision}, 4217--4229.

\bibitem[{Chen et~al.(2023)Chen, Chen, Jiao, and Jia}]{chen2023fantasia3d}
Chen, R.; Chen, Y.; Jiao, N.; and Jia, K. 2023.
\newblock Fantasia3d: Disentangling geometry and appearance for high-quality text-to-3d content creation.
\newblock In \emph{Proceedings of the IEEE/CVF International Conference on Computer Vision}, 22246--22256.

\bibitem[{Chen, Deng, and Wang(2023)}]{chen2023mimic3d}
Chen, X.; Deng, Y.; and Wang, B. 2023.
\newblock Mimic3d: Thriving 3d-aware gans via 3d-to-2d imitation.
\newblock In \emph{2023 IEEE/CVF International Conference on Computer Vision (ICCV)}, 2338--2348. IEEE Computer Society.

\bibitem[{Deng et~al.(2019)Deng, Guo, Xue, and Zafeiriou}]{deng2019arcface}
Deng, J.; Guo, J.; Xue, N.; and Zafeiriou, S. 2019.
\newblock Arcface: Additive angular margin loss for deep face recognition.
\newblock In \emph{Proceedings of the IEEE/CVF conference on computer vision and pattern recognition}, 4690--4699.

\bibitem[{Eftekhar et~al.(2021)Eftekhar, Sax, Malik, and Zamir}]{eftekhar2021omnidata}
Eftekhar, A.; Sax, A.; Malik, J.; and Zamir, A. 2021.
\newblock Omnidata: A scalable pipeline for making multi-task mid-level vision datasets from 3d scans.
\newblock In \emph{Proceedings of the IEEE/CVF International Conference on Computer Vision}, 10786--10796.

\bibitem[{Feng et~al.(2021)Feng, Feng, Black, and Bolkart}]{DECA:Siggraph2021}
Feng, Y.; Feng, H.; Black, M.~J.; and Bolkart, T. 2021.
\newblock Learning an Animatable Detailed {3D} Face Model from In-The-Wild Images.

\bibitem[{Gafni et~al.(2021)Gafni, Thies, Zollhofer, and Nie{\ss}ner}]{gafni2021dynamic}
Gafni, G.; Thies, J.; Zollhofer, M.; and Nie{\ss}ner, M. 2021.
\newblock Dynamic neural radiance fields for monocular 4d facial avatar reconstruction.
\newblock In \emph{Proceedings of the IEEE/CVF Conference on Computer Vision and Pattern Recognition}, 8649--8658.

\bibitem[{Grassal et~al.(2022)Grassal, Prinzler, Leistner, Rother, Nie{\ss}ner, and Thies}]{grassal2022neural}
Grassal, P.-W.; Prinzler, M.; Leistner, T.; Rother, C.; Nie{\ss}ner, M.; and Thies, J. 2022.
\newblock Neural head avatars from monocular rgb videos.
\newblock In \emph{Proceedings of the IEEE/CVF Conference on Computer Vision and Pattern Recognition}, 18653--18664.

\bibitem[{Guo et~al.(2020)Guo, Zhu, Yang, Yang, Lei, and Li}]{guo2020towards}
Guo, J.; Zhu, X.; Yang, Y.; Yang, F.; Lei, Z.; and Li, S.~Z. 2020.
\newblock Towards Fast, Accurate and Stable 3D Dense Face Alignment.
\newblock In \emph{Proceedings of the European Conference on Computer Vision (ECCV)}.

\bibitem[{Guo et~al.(2023)Guo, Liu, Shao, Laforte, Voleti, Luo, Chen, Zou, Wang, Cao, and Zhang}]{threestudio2023}
Guo, Y.-C.; Liu, Y.-T.; Shao, R.; Laforte, C.; Voleti, V.; Luo, G.; Chen, C.-H.; Zou, Z.-X.; Wang, C.; Cao, Y.-P.; and Zhang, S.-H. 2023.
\newblock threestudio: A unified framework for 3D content generation.
\newblock \url{https://github.com/threestudio-project/threestudio}.

\bibitem[{Han et~al.(2024)Han, Cao, Han, Zhu, Deng, Song, Xiang, and Wong}]{han2024headsculpt}
Han, X.; Cao, Y.; Han, K.; Zhu, X.; Deng, J.; Song, Y.-Z.; Xiang, T.; and Wong, K.-Y.~K. 2024.
\newblock Headsculpt: Crafting 3d head avatars with text.
\newblock \emph{Advances in Neural Information Processing Systems}, 36.

\bibitem[{Han et~al.(2023)Han, Zhang, Zhu, Li, Ge, Li, Wang, Liu, Liu, and Tai}]{han2023generalist}
Han, Y.; Zhang, J.; Zhu, J.; Li, X.; Ge, Y.; Li, W.; Wang, C.; Liu, Y.; Liu, X.; and Tai, Y. 2023.
\newblock A Generalist FaceX via Learning Unified Facial Representation.
\newblock \emph{arXiv preprint arXiv:2401.00551}.

\bibitem[{Huang et~al.(2023)Huang, Shao, Zhang, Zhang, Feng, Liu, and Wang}]{huang2023humannorm}
Huang, X.; Shao, R.; Zhang, Q.; Zhang, H.; Feng, Y.; Liu, Y.; and Wang, Q. 2023.
\newblock Humannorm: Learning normal diffusion model for high-quality and realistic 3d human generation.
\newblock \emph{arXiv preprint arXiv:2310.01406}.

\bibitem[{Karras, Laine, and Aila(2019)}]{karras2019style}
Karras, T.; Laine, S.; and Aila, T. 2019.
\newblock A style-based generator architecture for generative adversarial networks.
\newblock In \emph{Proceedings of the IEEE/CVF conference on computer vision and pattern recognition}, 4401--4410.

\bibitem[{Kirschstein et~al.(2023)Kirschstein, Qian, Giebenhain, Walter, and Nie{\ss}ner}]{kirschstein2023nersemble}
Kirschstein, T.; Qian, S.; Giebenhain, S.; Walter, T.; and Nie{\ss}ner, M. 2023.
\newblock Nersemble: Multi-view radiance field reconstruction of human heads.
\newblock \emph{ACM Transactions on Graphics (TOG)}, 42(4): 1--14.

\bibitem[{Li et~al.(2023)Li, Li, Savarese, and Hoi}]{li2023blip}
Li, J.; Li, D.; Savarese, S.; and Hoi, S. 2023.
\newblock Blip-2: Bootstrapping language-image pre-training with frozen image encoders and large language models.
\newblock In \emph{International conference on machine learning}, 19730--19742. PMLR.

\bibitem[{Lin et~al.(2023)Lin, Gao, Tang, Takikawa, Zeng, Huang, Kreis, Fidler, Liu, and Lin}]{lin2023magic3d}
Lin, C.-H.; Gao, J.; Tang, L.; Takikawa, T.; Zeng, X.; Huang, X.; Kreis, K.; Fidler, S.; Liu, M.-Y.; and Lin, T.-Y. 2023.
\newblock Magic3d: High-resolution text-to-3d content creation.
\newblock In \emph{Proceedings of the IEEE/CVF Conference on Computer Vision and Pattern Recognition}, 300--309.

\bibitem[{Liu et~al.(2023{\natexlab{a}})Liu, Wang, Wan, Shen, Song, Liao, and Chen}]{liu2023headartist}
Liu, H.; Wang, X.; Wan, Z.; Shen, Y.; Song, Y.; Liao, J.; and Chen, Q. 2023{\natexlab{a}}.
\newblock HeadArtist: Text-conditioned 3D Head Generation with Self Score Distillation.
\newblock \emph{arXiv preprint arXiv:2312.07539}.

\bibitem[{Liu et~al.(2023{\natexlab{b}})Liu, Wu, Van~Hoorick, Tokmakov, Zakharov, and Vondrick}]{liu2023zero}
Liu, R.; Wu, R.; Van~Hoorick, B.; Tokmakov, P.; Zakharov, S.; and Vondrick, C. 2023{\natexlab{b}}.
\newblock Zero-1-to-3: Zero-shot one image to 3d object.
\newblock In \emph{Proceedings of the IEEE/CVF International Conference on Computer Vision}, 9298--9309.

\bibitem[{Liu et~al.(2023{\natexlab{c}})Liu, Lin, Zeng, Long, Liu, Komura, and Wang}]{liu2023syncdreamer}
Liu, Y.; Lin, C.; Zeng, Z.; Long, X.; Liu, L.; Komura, T.; and Wang, W. 2023{\natexlab{c}}.
\newblock Syncdreamer: Generating multiview-consistent images from a single-view image.
\newblock \emph{arXiv preprint arXiv:2309.03453}.

\bibitem[{Long et~al.(2023)Long, Guo, Lin, Liu, Dou, Liu, Ma, Zhang, Habermann, Theobalt et~al.}]{long2023wonder3d}
Long, X.; Guo, Y.-C.; Lin, C.; Liu, Y.; Dou, Z.; Liu, L.; Ma, Y.; Zhang, S.-H.; Habermann, M.; Theobalt, C.; et~al. 2023.
\newblock Wonder3d: Single image to 3d using cross-domain diffusion.
\newblock \emph{arXiv preprint arXiv:2310.15008}.

\bibitem[{Melas-Kyriazi et~al.(2023)Melas-Kyriazi, Laina, Rupprecht, and Vedaldi}]{melas2023realfusion}
Melas-Kyriazi, L.; Laina, I.; Rupprecht, C.; and Vedaldi, A. 2023.
\newblock Realfusion: 360deg reconstruction of any object from a single image.
\newblock In \emph{Proceedings of the IEEE/CVF conference on computer vision and pattern recognition}, 8446--8455.

\bibitem[{Munkberg et~al.(2022)Munkberg, Hasselgren, Shen, Gao, Chen, Evans, M{\"u}ller, and Fidler}]{munkberg2022extracting}
Munkberg, J.; Hasselgren, J.; Shen, T.; Gao, J.; Chen, W.; Evans, A.; M{\"u}ller, T.; and Fidler, S. 2022.
\newblock Extracting triangular 3d models, materials, and lighting from images.
\newblock In \emph{Proceedings of the IEEE/CVF Conference on Computer Vision and Pattern Recognition}, 8280--8290.

\bibitem[{Pei et~al.(2024)Pei, Zhang, Hu, Zhai, Wang, Zhang, Yang, Shen, and Tao}]{pei2024deepfake}
Pei, G.; Zhang, J.; Hu, M.; Zhai, G.; Wang, C.; Zhang, Z.; Yang, J.; Shen, C.; and Tao, D. 2024.
\newblock Deepfake Generation and Detection: A Benchmark and Survey.
\newblock \emph{arXiv preprint arXiv:2403.17881}.

\bibitem[{Poole et~al.(2022)Poole, Jain, Barron, and Mildenhall}]{poole2022dreamfusion}
Poole, B.; Jain, A.; Barron, J.~T.; and Mildenhall, B. 2022.
\newblock Dreamfusion: Text-to-3d using 2d diffusion.
\newblock \emph{arXiv preprint arXiv:2209.14988}.

\bibitem[{Qian et~al.(2023)Qian, Mai, Hamdi, Ren, Siarohin, Li, Lee, Skorokhodov, Wonka, Tulyakov et~al.}]{qian2023magic123}
Qian, G.; Mai, J.; Hamdi, A.; Ren, J.; Siarohin, A.; Li, B.; Lee, H.-Y.; Skorokhodov, I.; Wonka, P.; Tulyakov, S.; et~al. 2023.
\newblock Magic123: One image to high-quality 3d object generation using both 2d and 3d diffusion priors.
\newblock \emph{arXiv preprint arXiv:2306.17843}.

\bibitem[{Ramesh et~al.(2022)Ramesh, Dhariwal, Nichol, Chu, and Chen}]{ramesh2022hierarchical}
Ramesh, A.; Dhariwal, P.; Nichol, A.; Chu, C.; and Chen, M. 2022.
\newblock Hierarchical text-conditional image generation with clip latents.
\newblock \emph{arXiv preprint arXiv:2204.06125}, 1(2): 3.

\bibitem[{Ramesh et~al.(2021)Ramesh, Pavlov, Goh, Gray, Voss, Radford, Chen, and Sutskever}]{ramesh2021zero}
Ramesh, A.; Pavlov, M.; Goh, G.; Gray, S.; Voss, C.; Radford, A.; Chen, M.; and Sutskever, I. 2021.
\newblock Zero-shot text-to-image generation.
\newblock In \emph{International conference on machine learning}, 8821--8831. Pmlr.

\bibitem[{Ranftl, Bochkovskiy, and Koltun(2021)}]{Ranftl2021}
Ranftl, R.; Bochkovskiy, A.; and Koltun, V. 2021.
\newblock Vision Transformers for Dense Prediction.
\newblock \emph{ArXiv preprint}.

\bibitem[{Ranftl et~al.(2020)Ranftl, Lasinger, Hafner, Schindler, and Koltun}]{Ranftl2020}
Ranftl, R.; Lasinger, K.; Hafner, D.; Schindler, K.; and Koltun, V. 2020.
\newblock Towards Robust Monocular Depth Estimation: Mixing Datasets for Zero-shot Cross-dataset Transfer.
\newblock \emph{IEEE Transactions on Pattern Analysis and Machine Intelligence (TPAMI)}.

\bibitem[{Roich et~al.(2022)Roich, Mokady, Bermano, and Cohen-Or}]{roich2022pivotal}
Roich, D.; Mokady, R.; Bermano, A.~H.; and Cohen-Or, D. 2022.
\newblock Pivotal tuning for latent-based editing of real images.
\newblock \emph{ACM Transactions on graphics (TOG)}, 42(1): 1--13.

\bibitem[{Rombach et~al.(2022)Rombach, Blattmann, Lorenz, Esser, and Ommer}]{rombach2022high}
Rombach, R.; Blattmann, A.; Lorenz, D.; Esser, P.; and Ommer, B. 2022.
\newblock High-resolution image synthesis with latent diffusion models.
\newblock In \emph{Proceedings of the IEEE/CVF conference on computer vision and pattern recognition}, 10684--10695.

\bibitem[{Saharia et~al.(2022)Saharia, Chan, Saxena, Li, Whang, Denton, Ghasemipour, Gontijo~Lopes, Karagol~Ayan, Salimans et~al.}]{saharia2022photorealistic}
Saharia, C.; Chan, W.; Saxena, S.; Li, L.; Whang, J.; Denton, E.~L.; Ghasemipour, K.; Gontijo~Lopes, R.; Karagol~Ayan, B.; Salimans, T.; et~al. 2022.
\newblock Photorealistic text-to-image diffusion models with deep language understanding.
\newblock \emph{Advances in neural information processing systems}, 35: 36479--36494.

\bibitem[{Sanghi et~al.(2022)Sanghi, Chu, Lambourne, Wang, Cheng, Fumero, and Malekshan}]{sanghi2022clip}
Sanghi, A.; Chu, H.; Lambourne, J.~G.; Wang, Y.; Cheng, C.-Y.; Fumero, M.; and Malekshan, K.~R. 2022.
\newblock Clip-forge: Towards zero-shot text-to-shape generation.
\newblock In \emph{Proceedings of the IEEE/CVF Conference on Computer Vision and Pattern Recognition}, 18603--18613.

\bibitem[{Shen et~al.(2021)Shen, Gao, Yin, Liu, and Fidler}]{shen2021dmtet}
Shen, T.; Gao, J.; Yin, K.; Liu, M.-Y.; and Fidler, S. 2021.
\newblock Deep Marching Tetrahedra: a Hybrid Representation for High-Resolution 3D Shape Synthesis.
\newblock In \emph{Advances in Neural Information Processing Systems (NeurIPS)}, 6087--6101.

\bibitem[{Sun et~al.(2023)Sun, Zhang, Shao, Wang, Liu, Xie, and Liu}]{sun2023dreamcraft3d}
Sun, J.; Zhang, B.; Shao, R.; Wang, L.; Liu, W.; Xie, Z.; and Liu, Y. 2023.
\newblock Dreamcraft3d: Hierarchical 3d generation with bootstrapped diffusion prior.
\newblock \emph{arXiv preprint arXiv:2310.16818}.

\bibitem[{Tang et~al.(2023)Tang, Wang, Zhang, Zhang, Yi, Ma, and Chen}]{tang2023make}
Tang, J.; Wang, T.; Zhang, B.; Zhang, T.; Yi, R.; Ma, L.; and Chen, D. 2023.
\newblock Make-it-3d: High-fidelity 3d creation from a single image with diffusion prior.
\newblock In \emph{Proceedings of the IEEE/CVF International Conference on Computer Vision}, 22819--22829.

\bibitem[{Wang et~al.(2022)Wang, Chai, He, Chen, and Liao}]{wang2022clip}
Wang, C.; Chai, M.; He, M.; Chen, D.; and Liao, J. 2022.
\newblock Clip-nerf: Text-and-image driven manipulation of neural radiance fields.
\newblock In \emph{Proceedings of the IEEE/CVF Conference on Computer Vision and Pattern Recognition}, 3835--3844.

\bibitem[{Wang et~al.(2023)Wang, Zhang, Zhang, Gu, Bao, Baltrusaitis, Shen, Chen, Wen, Chen et~al.}]{wang2023rodin}
Wang, T.; Zhang, B.; Zhang, T.; Gu, S.; Bao, J.; Baltrusaitis, T.; Shen, J.; Chen, D.; Wen, F.; Chen, Q.; et~al. 2023.
\newblock Rodin: A generative model for sculpting 3d digital avatars using diffusion.
\newblock In \emph{Proceedings of the IEEE/CVF conference on computer vision and pattern recognition}, 4563--4573.

\bibitem[{Wang et~al.(2024)Wang, Lu, Wang, Bao, Li, Su, and Zhu}]{wang2024prolificdreamer}
Wang, Z.; Lu, C.; Wang, Y.; Bao, F.; Li, C.; Su, H.; and Zhu, J. 2024.
\newblock Prolificdreamer: High-fidelity and diverse text-to-3d generation with variational score distillation.
\newblock \emph{Advances in Neural Information Processing Systems}, 36.

\bibitem[{Xu et~al.(2023)Xu, Chen, Li, Zhang, Wang, Zheng, and Liu}]{xu2023gaussian}
Xu, Y.; Chen, B.; Li, Z.; Zhang, H.; Wang, L.; Zheng, Z.; and Liu, Y. 2023.
\newblock Gaussian head avatar: Ultra high-fidelity head avatar via dynamic gaussians.
\newblock \emph{arXiv preprint arXiv:2312.03029}.

\bibitem[{Yariv et~al.(2021)Yariv, Gu, Kasten, and Lipman}]{yariv2021volume}
Yariv, L.; Gu, J.; Kasten, Y.; and Lipman, Y. 2021.
\newblock Volume rendering of neural implicit surfaces.
\newblock \emph{Advances in Neural Information Processing Systems}, 34: 4805--4815.

\bibitem[{Ye et~al.(2023)Ye, Zhang, Liu, Han, and Yang}]{ye2023ip}
Ye, H.; Zhang, J.; Liu, S.; Han, X.; and Yang, W. 2023.
\newblock Ip-adapter: Text compatible image prompt adapter for text-to-image diffusion models.
\newblock \emph{arXiv preprint arXiv:2308.06721}.

\bibitem[{Yu et~al.(2023)Yu, Yuan, Cao, Gao, Li, Quan, Shan, and Tian}]{yu2023hifi}
Yu, W.; Yuan, L.; Cao, Y.-P.; Gao, X.; Li, X.; Quan, L.; Shan, Y.; and Tian, Y. 2023.
\newblock Hifi-123: Towards high-fidelity one image to 3d content generation.
\newblock \emph{arXiv preprint arXiv:2310.06744}.

\bibitem[{Zhang, Rao, and Agrawala(2023)}]{zhang2023adding}
Zhang, L.; Rao, A.; and Agrawala, M. 2023.
\newblock Adding conditional control to text-to-image diffusion models.
\newblock In \emph{Proceedings of the IEEE/CVF International Conference on Computer Vision}, 3836--3847.

\bibitem[{Zhang et~al.(2018)Zhang, Isola, Efros, Shechtman, and Wang}]{zhang2018unreasonable}
Zhang, R.; Isola, P.; Efros, A.~A.; Shechtman, E.; and Wang, O. 2018.
\newblock The unreasonable effectiveness of deep features as a perceptual metric.
\newblock In \emph{Proceedings of the IEEE conference on computer vision and pattern recognition}, 586--595.

\bibitem[{Zheng et~al.(2023{\natexlab{a}})Zheng, Zhang, Yang, and Huang}]{Zheng_2023_CVPR}
Zheng, M.; Zhang, H.; Yang, H.; and Huang, D. 2023{\natexlab{a}}.
\newblock NeuFace: Realistic 3D Neural Face Rendering From Multi-View Images.
\newblock In \emph{Proceedings of the IEEE/CVF Conference on Computer Vision and Pattern Recognition (CVPR)}, 16868--16877.

\bibitem[{Zheng et~al.(2023{\natexlab{b}})Zheng, Yifan, Wetzstein, Black, and Hilliges}]{zheng2023pointavatar}
Zheng, Y.; Yifan, W.; Wetzstein, G.; Black, M.~J.; and Hilliges, O. 2023{\natexlab{b}}.
\newblock Pointavatar: Deformable point-based head avatars from videos.
\newblock In \emph{Proceedings of the IEEE/CVF conference on computer vision and pattern recognition}, 21057--21067.

\end{thebibliography}

\appendix
\section{Appendix Overview}
In the appendix, we will show more content not mentioned in the main paper, including: 
\begin{itemize}
    \item More implementation details in our method.
    \item More ablation study.
    \item More visual results.
    \item More comparison results with previous methods.
\end{itemize}

\section{More implementation details}

In the geometry and texture generation stage, we set the resolution of DMTet to 512 to achieve better geometric details. Additionally, the size of the unfolded texture map from the mesh is set to 1024x1024.

\subsection{Align landmark with initialized geometry}
In the geometry sculpting stage, landmark can constrain the face layout in the generation process~\cite{han2024headsculpt, liu2023headartist}, which effectively solves the multi-face Janus issue. Therefore, we need to obtain landmarks that are aligned with the initialized mesh to guide the geometry generation process. 

\begin{figure}[htb]
  \includegraphics[width=0.4\textwidth]{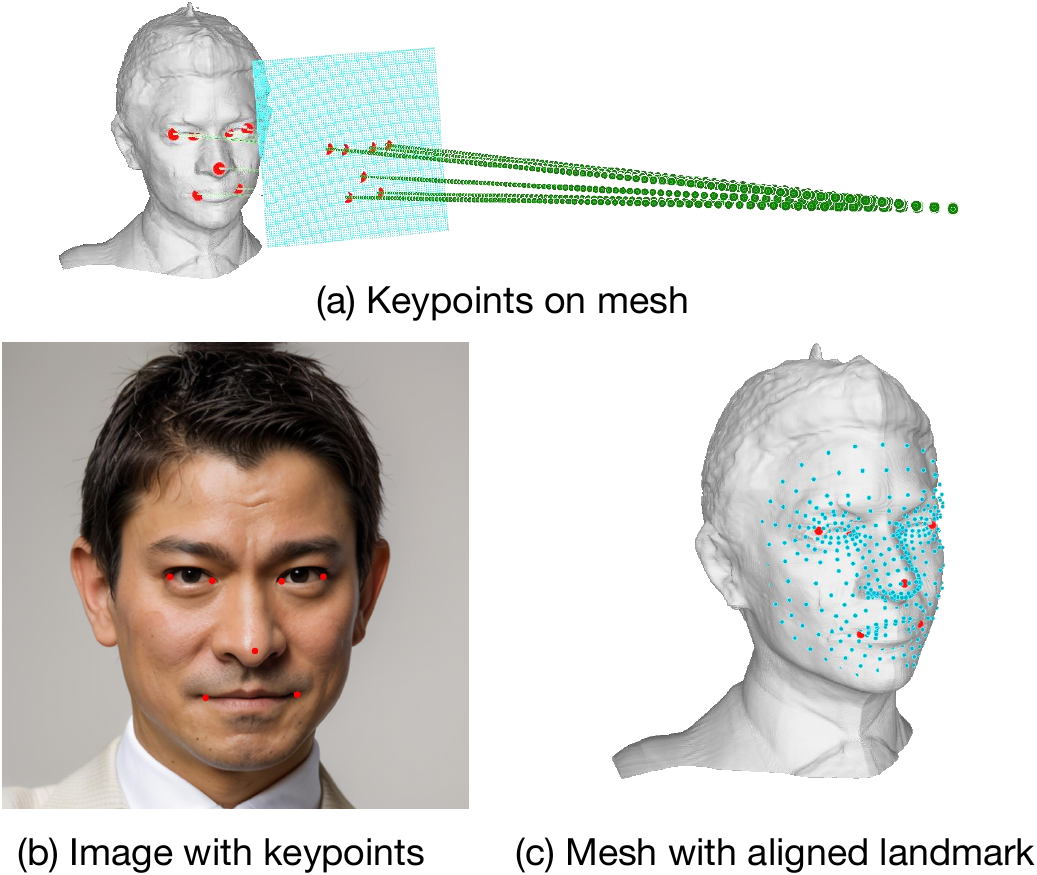}
  \caption{We compute the transformation matrix between source keypoints and target keypoints to align the landmark to the initialized mesh.}
  \label{fig:align}
\end{figure}

We first use face detector (3DDFA~\cite{guo2020towards}) to obtain 3D landmark of face from input image and get seven source keypoints as shown in Figure~\ref{fig:align} (b). At the same time, the predicted camera extrinsic parameters are obtained, and this extrinsic component is aligned with the camera parameters in 3D GAN~\cite{an2023panohead}. Next, utilizing the camera position, the imaging plane (blue plane), and the 3D coordinates of the key points on the imaging plane (red points on the blue plane), along with the initialized mesh, we employ the camera position as the origin of the rays (green rays). The directions of the rays are determined by the vectors from the camera position to the keypoints in the imaging plane. We use the Octree-accelerated ray-grid intersection method to compute the intersections of these rays with the initial mesh to obtain the target keypoints, as shown in Figure~\ref{fig:align} (a). We then calculate the transformation matrix from the source keypoints to the target points to align the landmarks to the initialized mesh, as shown in Figure~\ref{fig:align} (c).

\subsection{Texture inpainting details}
In the Progressive Texture Inpainting process, we set the camera trajectory to start from the front reference view, symmetrically select the viewpoint for inpainting, and gradually move the camera viewpoint to the back of the head. Specifically, the camera's horizontal angle moves according to $[0^\circ, 45^\circ, -45^\circ, 90^\circ, -90^\circ, 135^\circ, -135^\circ, 180^\circ]$, and the vertical angle is set to $-15^\circ$. In this way, we can generate a coherent head texture.

\begin{figure}[ht]
  \includegraphics[width=0.45\textwidth]{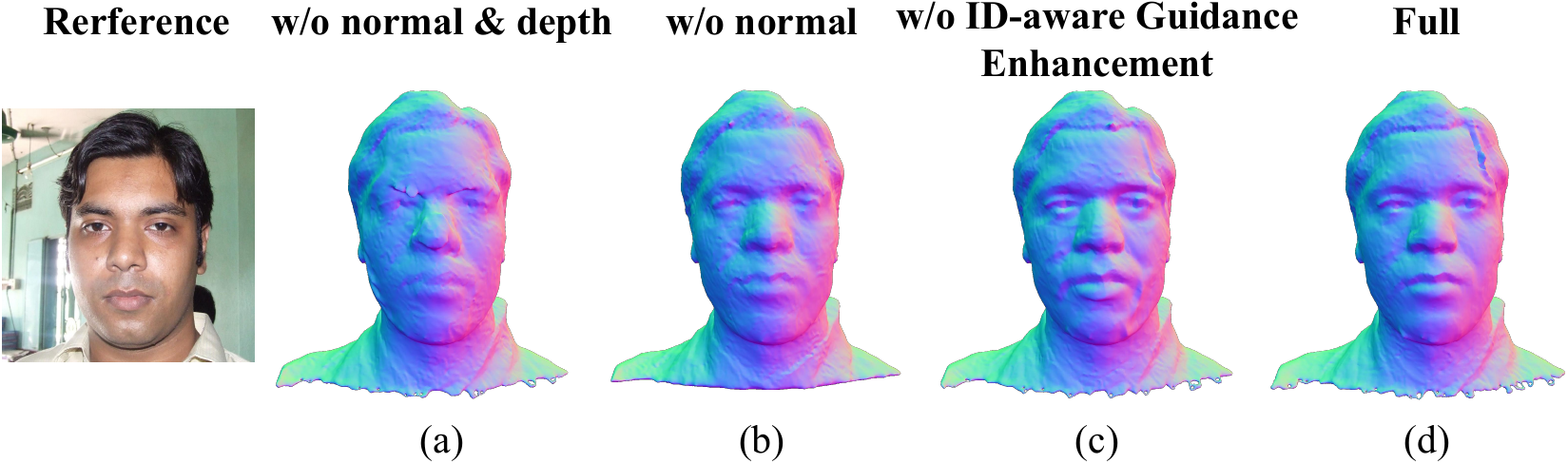}
  \caption{Visual result of ablation study. (a) Replace our ID-aware depth and normal with depth and normal predicted by DPT, (b) Geometry with only our ID-aware depth supervision, (c) Replace our ISD loss with SDS loss, (d) Our full method.}
  \label{fig:abla}
\end{figure}

\subsection{Texture refinement details}
In the texture refinement phase, in order to maintain the basic information of the initial texture and to eliminate the artifacts generated by inpainting, we add noise to the rendered image with a small timestep $t=120$ and denoise the noised image with ID diffusion to generate a higher quality head texture. We employ the random camera sampling strategy with an elevation range of $[-20^\circ, 45^\circ]$, an azimuth range of $[-180^\circ, 180^\circ]$, a fovy range of $[-30^\circ, 45^\circ]$, and a camera distance range of [2.5, 4.0].

\section{More ablation}
The ablation result of our method in the geometry sculpting stage in Figure~\ref{fig:abla}. (a) Without both ID-aware depth and normal supervision in reference view, the lack of depth results in a flatter head frontal geometry, and a flaky artifact in the geometry, (b) The absence of ID-aware normal map leaves the geometry lacking in detail, (c) Without employing the ID-aware Guidance Enhancement, uneven geometry may arise in the reference viewpoint. This is due to the inconsistency between the geometric supervision under the reference viewpoint and the guidance provided by the SDS. (d) Our full model can generate high-quality geometry compared to (a), (b) and (c).

Quantitative ablation result in Table~\ref{tab:quan_abla} shows that ID-aware Initialization and Guidance greatly
enhance the generation quality and identity preservation ability.

\begin{table}[]
\centering
\small
\setlength{\tabcolsep}{4pt} 
\caption{Quantitative ablations on head generation task. We compute LPIPS under the reference view, ClIP-I and ID score under novel views.}
\label{tab:quan_abla}
\begin{tabular}{l|lll}
\hline
\textbf{}                          & \textbf{LPIPS$\downarrow$} & \textbf{CLIP-I$\uparrow$} & \textbf{ID$\uparrow$}    \\ \hline
w/o ID-aware Initialization        & 0.218           & 0.5790           & 0.2296          \\
w/o ID-aware Guidance              & 0.058           & 0.6490           & 0.2984          \\
w/o ID-aware Geo. Supervision  & 0.050           & 0.6143           & 0.4337          \\
w/o Texture Refinement             & 0.074           & 0.6368           & 0.3965          \\
w/o Progressive Tex. Inpainting & 0.059           & 0.6478           & 0.4269          \\
Full model (ours)                  & \textbf{0.039}  & \textbf{0.6605}  & \textbf{0.4451} \\ \hline
\end{tabular}
\end{table}

\section{More result}
\begin{figure*}[b]
  \centering
  \includegraphics[width=0.95\textwidth]{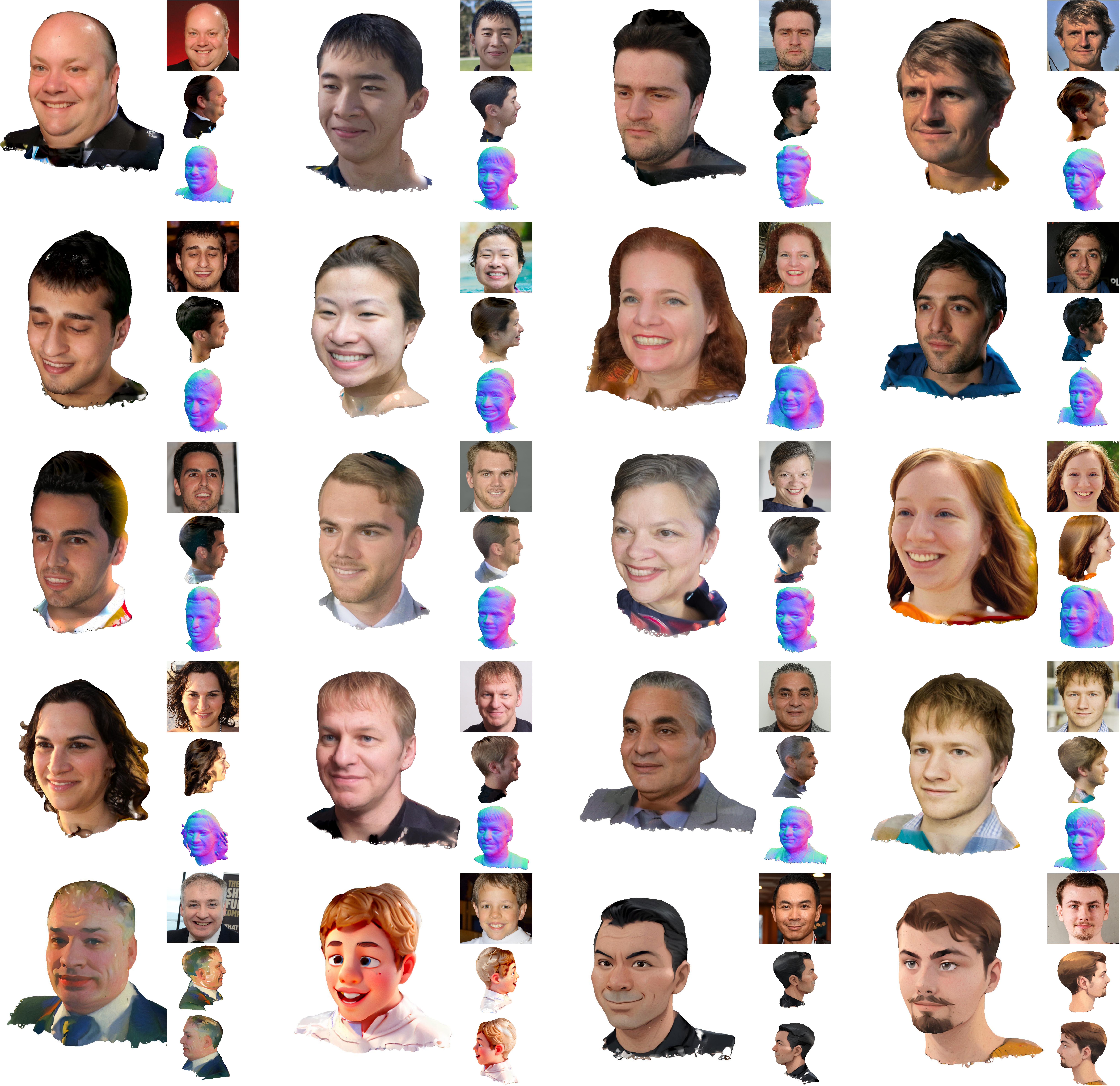}
  \caption{More visual result. Our method has high-quality geometry and photorealistic textures. Simultaneously, the generated 3D head models exhibit a high degree of consistency with the reference portrait images, showcasing the superior performance of our approach.}
  \label{fig:result}
\end{figure*}

Figure~\ref{fig:result} demonstrates the outstanding performance of our method in generating the 3D head from a single in-the-wild portrait image. Our results exhibit high-quality geometry and photorealistic textures. Furthermore, by employing diverse stylized diffusion models, we are capable of generating stylized 3D heads while preserving the portrait's identity.

\textbf{{Advantages over Panohead.}} 
ID-Sculpt surpasses Panohead in 3 aspects (shown in the Figure below): (1) Better geometry quality (eyes and mouth region). (2) Better texture quality (eye region). (3) Better 3D-consistency: Panohead has weak 3D consistency due to the 2D super-resolution module. We use spatiotemporal textures~\cite{chen2023mimic3d} to visualize 3D-consistency, which is obtained by stacking the pixels of a fixed line segment under continuous camera change, where smoothly tilted strips indicate better 3D consistency. Our spatiotemporal textures have fewer noisy patterns.

\begin{figure}[ht]
  \centering
  \includegraphics[width=0.85\linewidth]{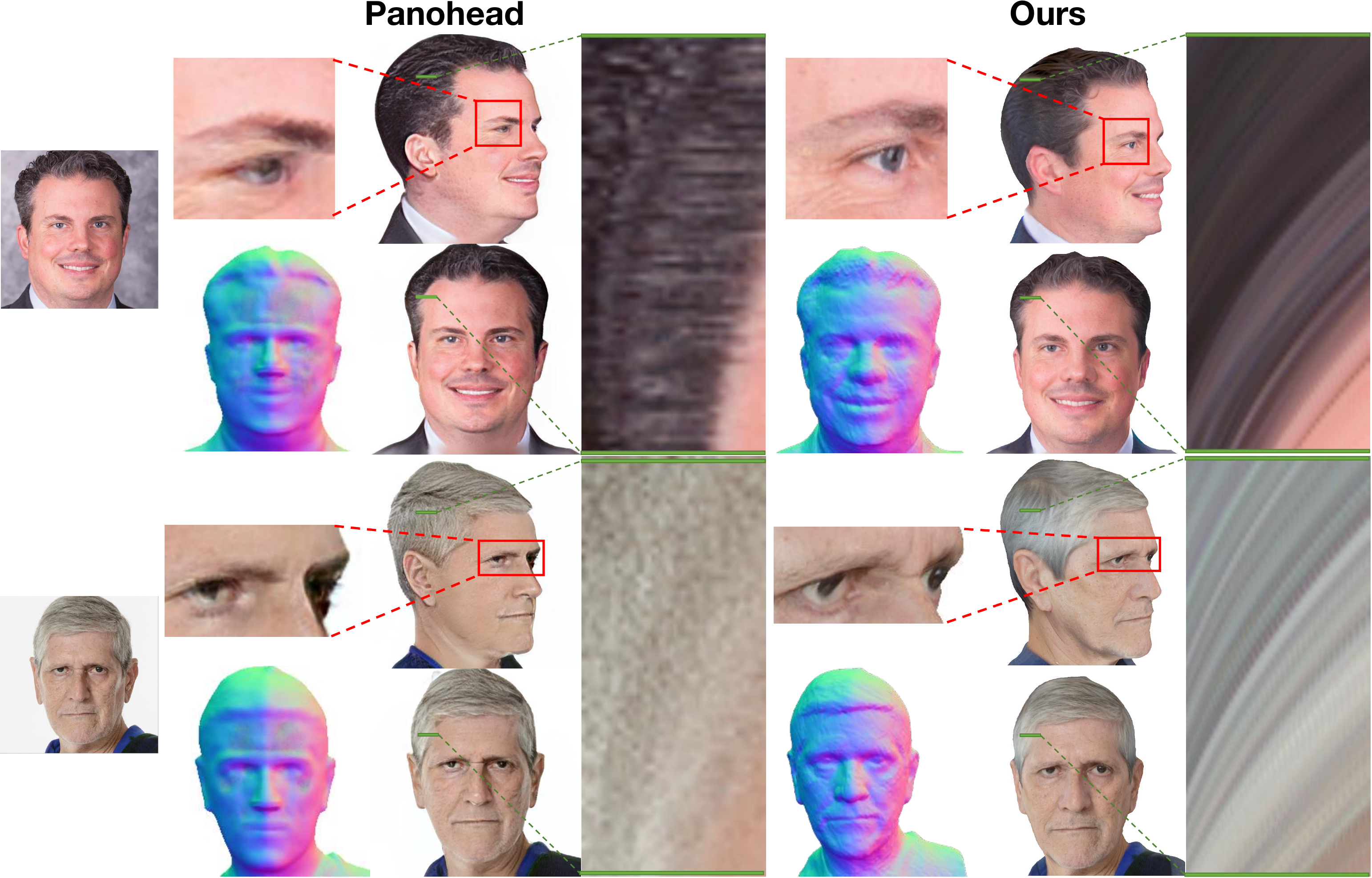}
  \caption{Comparasion between Panohead and our method.}
  \label{fig:consistency}
\end{figure}

\section{More comparison}
\begin{figure*}[t]
  \includegraphics[width=0.95\textwidth]{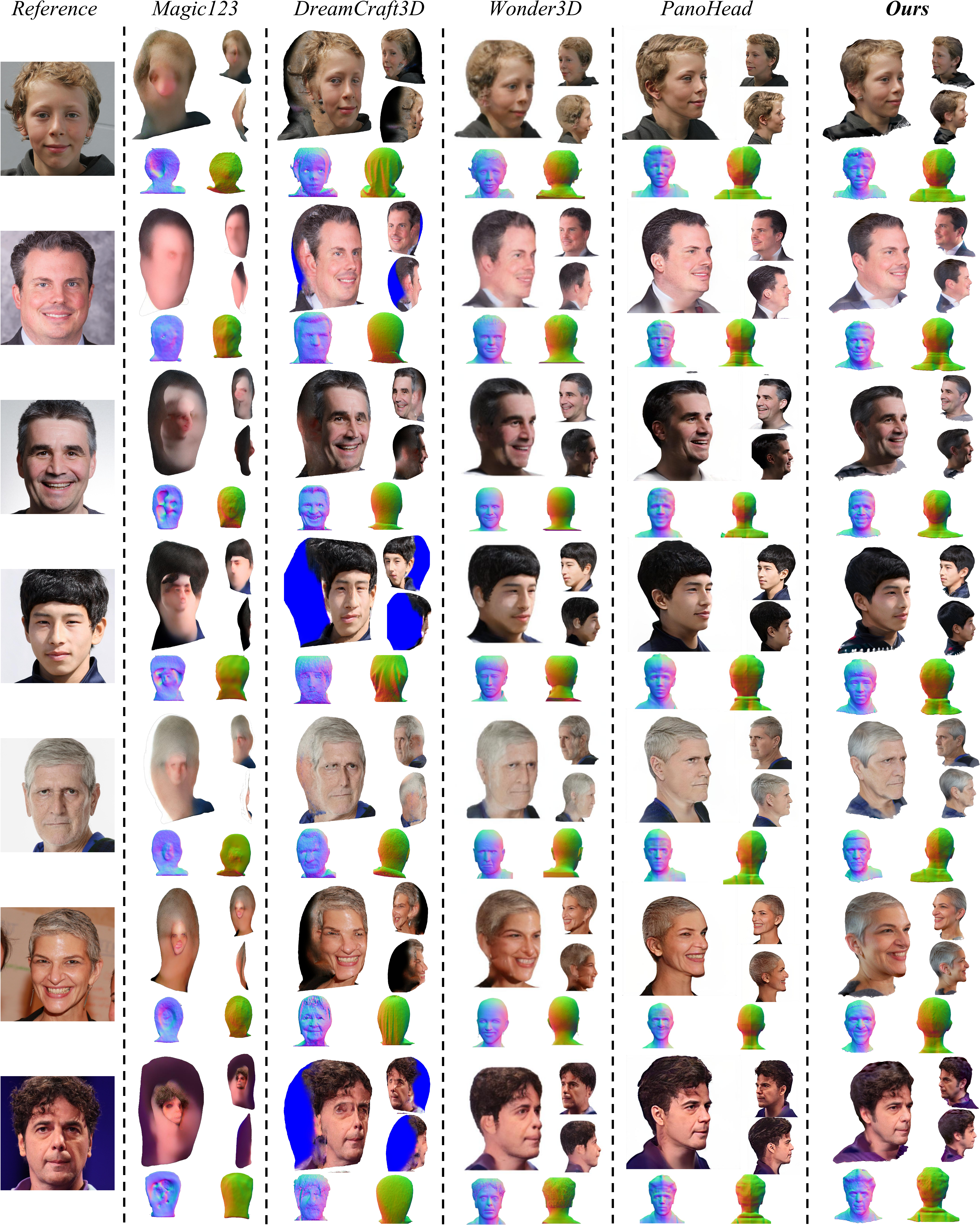}
  \caption{More comparison result. Compared to previous methods, our approach possesses more reasonable and richer geometry details, while maintaining photorealistic textures in both side and back views.}
  \label{fig:compare}
\end{figure*}

We show more comparison of our method with previous methods in Figure~\ref{fig:compare}. Compared with image-to-3D methods (Magic123~\cite{qian2023magic123}, Dreamcraft3D~\cite{sun2023dreamcraft3d}, wonder3D~\cite{long2023wonder3d}), our method has better geometric and texture quality in side view and back view. Also, our method has better facial geometric details compared to Panohead~\cite{an2023panohead}.

\end{document}